\documentclass[11pt,a4paper]{article}
\usepackage[margin=1in]{geometry}  

\usepackage{graphicx}
\usepackage{booktabs}
\usepackage{soul}
\usepackage{caption}
\usepackage{array}
\usepackage[table,x11names]{xcolor}
\usepackage{multirow}
\usepackage{calc}
\usepackage{amsmath}
\usepackage{amssymb}
\usepackage[accsupp]{axessibility}

\usepackage{orcidlink}
\usepackage[most]{tcolorbox}
\usepackage{algorithm}
\usepackage{algorithmic}
\usepackage{subcaption}
\usepackage{float}
\usepackage{enumitem}
\usepackage{adjustbox}
\usepackage{makecell}
\usepackage{tabularx}
\usepackage{ragged2e}

\usepackage{hyperref}
\hypersetup{
    colorlinks=true,
    linkcolor=blue,
    citecolor=blue,
    urlcolor=blue
}

\begin{document}

\title{SpecEdit: Training-Free Acceleration for Diffusion based Image Editing via Semantic Locking}

\author{
Zhengan Yan$^1$$^*$ \quad Shikang Zheng$^1$$^*$ \quad Haoran Qin$^1$$^,$$^3$$^*$ \quad Xiaobing Tu$^2$ \quad Yinggui Wang$^2$ \\
Jiacheng Liu$^{1}$ \quad Jiaxuan Ren$^1$$^,$$^4$ \quad Yuqi Lin$^1$$^,$$^5$ \quad Peiliang Cai$^1$ \quad Jinkui Ren$^2$ \\
Xiantao Zhang$^2$ \quad Linfeng Zhang$^1$$^\dagger$ \\[6pt]
$^1$ EPIC Lab , Shanghai Jiao Tong University \quad $^2$ Alibaba Group \quad $^3$ Shandong University \\
$^4$ UESTC \quad $^5$ Jilin University \\
}
\makeatletter
\renewcommand{\thefootnote}{}  
\makeatother

\date{}
\maketitle

\footnotetext[1]{$^*$ Equal contribution.}
\footnotetext[2]{$^\dagger$ Corresponding author.}
\begin{abstract}
Diffusion-based image editing offers strong semantic controllability, but remains computationally expensive due to iterative high-resolution denoising over all spatial tokens. Dynamic-resolution sampling reduces this cost by performing early steps at reduced resolution. However, existing approaches prioritize upsampling using low-level heuristics such as edge detection or channel variance, which are weakly aligned with editing semantics and may lead to structural inconsistency. 
Moreover, spatial regions are often upsampled without verifying whether semantic modification is actually required, resulting in redundant high-resolution computation and accumulated errors.
Therefore, we propose \textbf{SpecEdit} (\textbf{Spec}ulative \textbf{Edit}ing), a training-free dynamic-resolution framework tailored for diffusion-based image editing. 
SpecEdit follows a draft-and-verify scheme: a low-resolution draft first estimates the semantic outcome, after which token-level discrepancies are used to identify edit-relevant tokens for high-resolution denoising, while the remaining tokens stay at a coarse resolution.
Experiments on Qwen-Image-Edit and FLUX.1-Kontext-dev demonstrate up to $10\times$ and $7\times$ acceleration, while maintaining strong quality. 
SpecEdit is complementary to step distillation and other acceleration techniques, achieving up to $13\times$ speedup when combined with existing methods. 
Our code is in supplementary material and will be released on GitHub.
\end{abstract}

\hspace{\parindent}\textbf{Keywords:} Image Editing, Diffusion Acceleration
\section{Introduction}
\label{sec:intro}

Diffusion Transformers (DiTs) have become a dominant backbone for high-fidelity image generation and are increasingly adopted in instruction-based image editing systems. By conditioning on textual instructions and reference images, recent editors such as Qwen-Image-Edit~\cite{wu2025qwen} and FLUX.1-Kontext-dev~\cite{labs2025flux1kontextflowmatching} enable precise and localized semantic manipulation while preserving global realism. However, despite their strong controllability and visual quality, practical deployment remains constrained by inference efficiency. Image editing typically requires dozens of iterative denoising steps, and at each step the Transformer performs high-resolution token interactions across the entire latent grid. As attention complexity scales quadratically with the number of spatial tokens, this uniform full-resolution computation incurs substantial latency and computational cost, limiting interactive and real-time applications.

\begin{figure*}[htbp]
  \centering
  \includegraphics [trim=20 10 20 0, clip,width=1\linewidth]{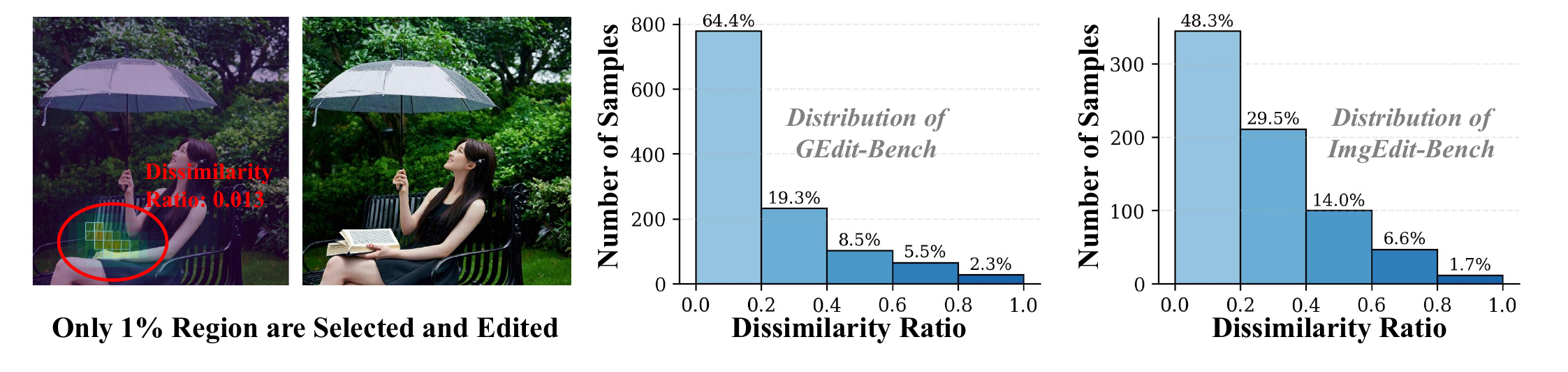}
  \caption{
  (\textbf{Left}) Visualization of the  semantic discrepancy estimated by our semantic verification mechanism.
  The discrepancy is computed between the  result and the original input in a perceptual feature space, revealing that only a small spatial region participates in the edit.
  (\textbf{Right}) Distribution of dissimilarity ratios on \textbf{GEdit-Bench}~\cite{liu2025step1x} and \textbf{ImgEdit-Bench}~\cite{ye2025imgedit}, where the ratio measures the proportion of tokens whose perceptual features change after editing.
  Most samples exhibit very low dissimilarity, indicating that real-world editing typically modifies only a small portion of the image.
  }
  \label{fig:data_summary}
\end{figure*}

Diffusion acceleration methods mainly explore two directions: temporal and spatial optimization. 
Temporal acceleration reduces the number of sampling steps using improved solvers, distillation, or timestep reduction strategies~\cite{liu2025reusing,liu2025timestep}. 
While effective for unconditional generation, aggressive step reduction can be fragile in editing scenarios, where identity preservation and structural consistency in unedited regions are critical. 
Semantic drift and localized artifacts become more likely when the denoising trajectory is shortened.

\begin{figure*}[htbp]
  \centering
  \includegraphics[trim=15 0 15 0, clip,width=1\linewidth]{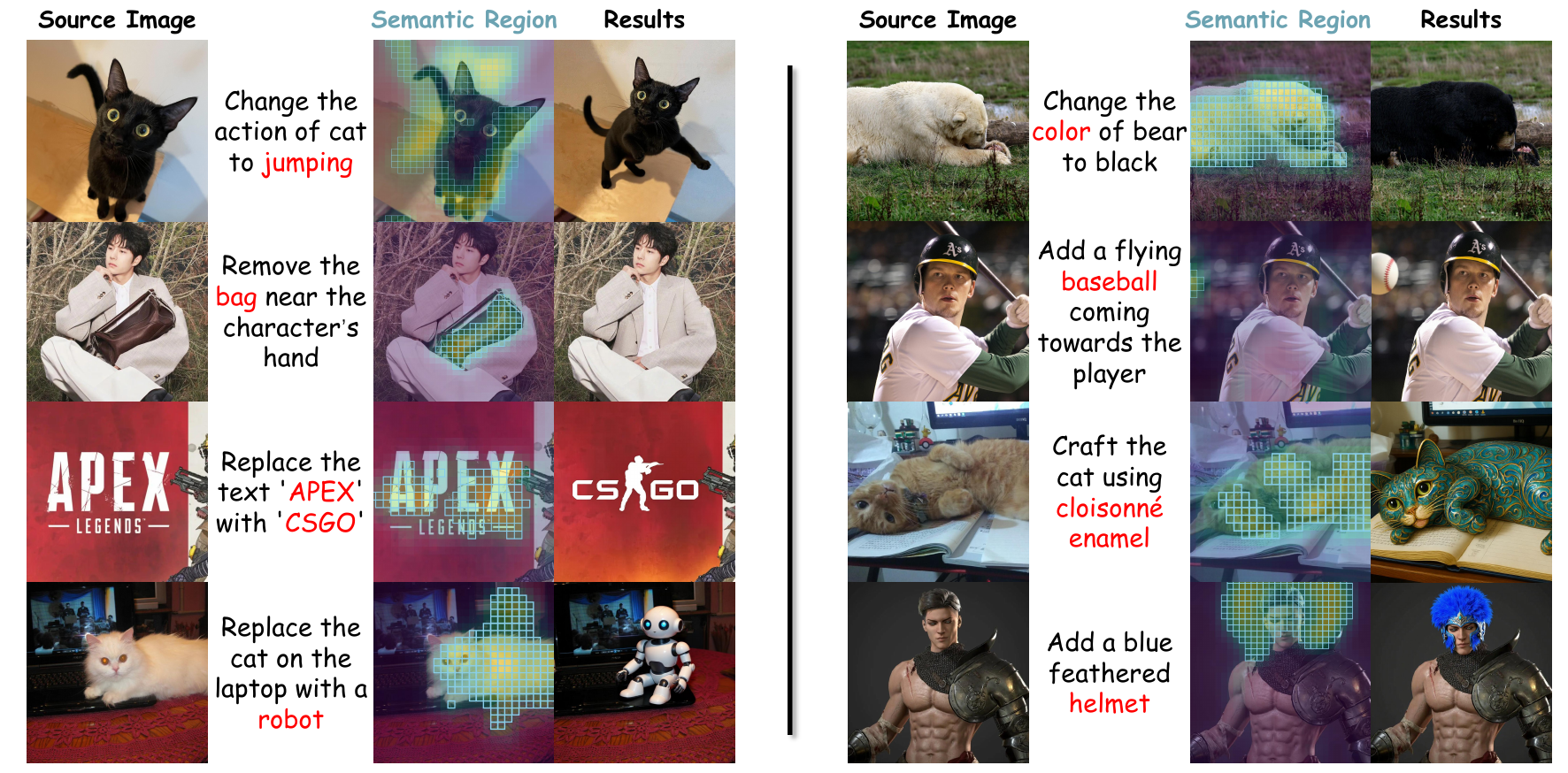}
  \caption{ Visualization of semantic locking captured by SpecEdit. }
  \label{fig:capture}
\end{figure*}

\begin{figure*}[h]
  \centering
  \includegraphics[width=\linewidth]{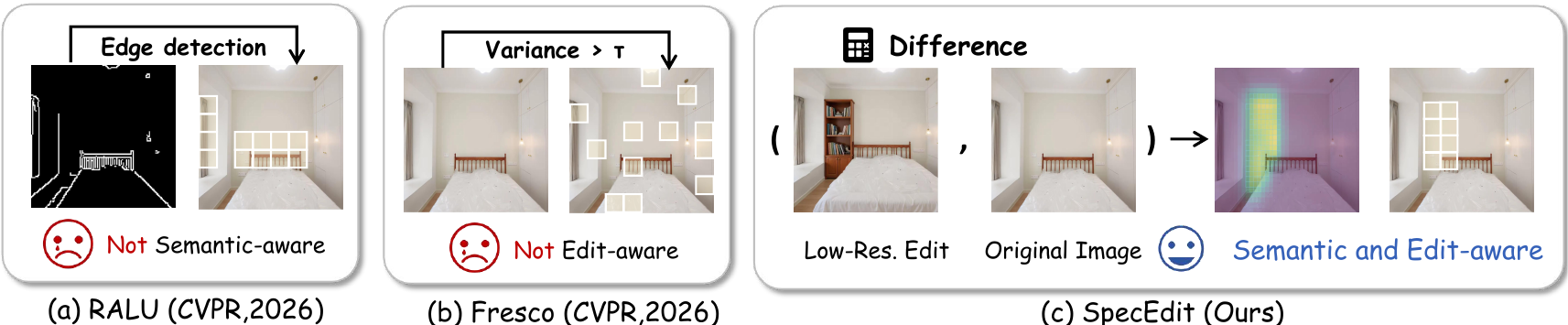}
  \caption{
  \textbf{Comparison with conventional dynamic-resolution methods.}
  (a) Edge-based methods (e.g., RALU~\cite{jeong2025upsample}) select refinement regions using edge responses, which mainly capture low-level structures and are not semantic-aware.
  (b) Channel-variance-based methods (e.g., Fresco~\cite{zheng2026sketch}) rely on channel statistics to guide region refinement but remain weakly aligned with editing semantics.
  (c) \textbf{SpecEdit (Ours)} generates an ultra-low-resolution draft and compares it with the original image to perform semantics-aware refinement, achieving stronger semantic alignment with the editing instruction.
  }
  \label{fig:intro_compare}
\end{figure*}

Spatial acceleration instead reduces per-step computation by lowering resolution or restricting token interactions~\cite{jeong2025upsample,tian2025training,zhang2025spargeattn}. 
This direction is particularly attractive for image editing because edits are inherently sparse: only a subset of spatial tokens undergo semantic modification. 
Applying high-resolution denoising uniformly to all tokens therefore results in substantial redundant computation.
This sparsity is empirically supported by benchmark statistics. 
As illustrated in Fig.~\ref{fig:data_summary}, semantic modification is spatially concentrated: only a subset of tokens exhibit substantial discrepancy under editing instructions, while most regions remain structurally stable. 
This observation is further confirmed statistically in Fig.~\ref{fig:data_summary}, where discrepant regions occupy only a limited fraction of the image on both GEdit and ImgEdit benchmarks.
These observations suggest that allocating high-resolution computation selectively to truly edited tokens could significantly improve efficiency without sacrificing fidelity. 
Dynamic-resolution sampling follows this intuition by performing early denoising at reduced resolution and selectively upsampling spatial tokens. 
However, directly adopting existing dynamic-resolution schemes for image editing remains nontrivial.

The central difficulty lies in determining which tokens deserve high-resolution computation. 
Most prior dynamic-resolution approaches are designed for unconditional or text-to-image generation and rely on low-level heuristics such as edge responses or feature variance to prioritize upsampling~\cite{jeong2025upsample,zheng2026sketch}. 
These signals are weakly correlated with editing intent: visually salient edges may correspond to background structures that should remain unchanged, whereas semantic edits often involve texture, attributes, or small objects with limited geometric saliency. 
As a result, heuristic-driven upsampling may allocate computation to irrelevant regions while neglecting tokens that genuinely require semantic modification. 
This mismatch leads to two issues in editing: degraded semantic consistency in modified regions and unnecessary high-resolution computation in unedited areas. As visualized in Fig.~\ref{fig:intro_compare}, heuristic-driven dynamic resolution often refines geometrically salient but semantically irrelevant regions, whereas SpecEdit explicitly locks semantically modified areas through draft-based verification.

In this work, we propose \textbf{SpecEdit} (\textbf{Spec}ulative \textbf{Edit}ing), a training-free dynamic-resolution framework tailored specifically for diffusion-based image editing. 
Our key insight is that high-resolution computation should be allocated based on \emph{semantic verification} rather than geometric heuristics. 
SpecEdit adopts a spatial draft and verification paradigm inspired by speculative decoding~\cite{leviathan2023fast}. 
Instead of performing full-resolution denoising over all tokens, we first generate a low-resolution draft in a heavily downsampled latent space, capturing the coarse semantic outcome of the edit at low cost. 
We then compute a token-level perceptual discrepancy map between the draft and the original input representation to identify edit-relevant tokens, effectively capturing the spatial regions that require semantic modification (Fig.~\ref{fig:capture}).

Only the verified tokens are priority-upsampled and fed into the Transformer for high-resolution denoising, while the remaining tokens stay at low resolution throughout the dynamic-resolution process and are upsampled only at the final reconstruction stage.
To stabilize global structure and prevent isolated artifacts, we additionally include a sparse set of uniformly sampled tokens in the upsampling set. 
This mixed-resolution design preserves global context while concentrating computation on semantically meaningful regions.
We evaluate SpecEdit on two representative editing models, Qwen-Image-Edit and FLUX.1-Kontext-dev, using the GEdit~\cite{liu2025step1x} and ImgEdit~\cite{ye2025imgedit} benchmarks. 
Across a wide range of computational budgets, SpecEdit improves the balance between efficiency and quality, achieving up to $10\times$ and $7\times$ acceleration, respectively, while maintaining strong semantic consistency and perceptual quality. 
SpecEdit is orthogonal to temporal acceleration and model compression techniques. When combined with distillation or caching, it can reach up to $13\times$ speedup.
Our contributions are summarized as follows:

\begin{itemize}[leftmargin=10pt,topsep=-2pt,label=\textbf{•}]
    \item \textbf{Dynamic resolution for image editing.}
    We are the first to systematically introduce dynamic-resolution sampling into diffusion-based image editing, adapting it to the unique requirement of semantic preservation under instruction-guided modification.

    \item \textbf{Semantics-aware draft-and-verify mechanism.}
    We propose \textbf{SpecEdit}, a training-free framework that employs a preliminary low-resolution draft for semantic discrepancy estimation and selectively expands verified tokens for high-resolution Transformer computation.

    \item \textbf{Outstanding performance.}
    Across Qwen-Image-Edit and FLUX.1-Kontext-dev, SpecEdit consistently improves acceleration without sacrificing semantic fidelity, and integrates seamlessly with temporal acceleration and model-efficiency techniques to provide additive speedups.
\end{itemize}

\section{Related Works}
While diffusion models have achieved remarkable performance in image generation, growing attention has been devoted to the field of image editing, with the pursuit of more accurate and efficient image editing approaches. Existing studies mainly focus on two aspects. One concerns semantic region distinction and the other focuses on model efficiency.

\noindent\textbf{Semantic Region Distinction in Image Editing.} Early researchers retained certain reference features by injecting external information (e.g., Contronet\cite{zhang2023adding} and KV injection techniques\cite{hertz2022prompt,kawar2023imagic}), and achieved promising results in both semantic information and image quality. With the further rapid development of transformers, diffusion transformer models can directly modify the corresponding semantic regions without manual masking. However, existing methods lack sufficient consideration for distinguishing edited and unedited regions, and fail to fully reflect the differences between the two during the denoising process. 

\noindent\textbf{Efficiency Optimization of Diffusion Models.} In terms of efficiency, researchers have carried out studies around the temporal and spatial dimensions, and proposed a series of optimization methods, which are divided into temporal acceleration and spatial acceleration:

\noindent\textbf{Temporal Acceleration.} The core idea of temporal acceleration is to reduce the number of timesteps as much as possible without degrading image quality. As the first method to propose a deterministic sampler, DDIM ~\cite{song2020denoising} enables fast image generation while controlling the perceptual loss at an extremely low level. In the aspect of high-order numerical solving, DPM-Solver\cite{lu2022dpm,lu2025dpm,zheng2023dpm} and its various variants take minimizing the local truncation error as the core means, achieving a precise balance between generation quality and efficiency. Progressive distillation technology\cite{salimans2022progressive} compresses the originally lengthy denoising chain into a shallow student model, reducing the model running cost from the training side. under the direct mapping  paradigm, consistency models abandon the traditional denoising logic and directly learn the fixed-point mapping relationship from noise to target signals, thereby further reducing the number of steps required in the inference stage. In addition, training-free computational optimization methods \cite{liu2025reusing,zheng2025letfeaturesdecidesolvers,liu2025freqca}without redundancy can reuse the hidden states and attention maps of different timesteps through the feature caching mechanism, thus significantly reducing the resource overhead caused by repeated computations.

\noindent\textbf{Spatial Acceleration.} Although sparse token interactions reduce the quadratic complexity of self-attention for diffusion acceleration, they yield merely limited practical speedup\cite{beltagy2020longformer,child2019generating,zaheer2020big}. Similarly, cascade diffusion models for coarse-to-fine synthesis introduce extra overhead from additional training pipelines and auxiliary networks\cite{ho2022cascaded,li2022srdiff}. State-of-the-art training-free dynamic resolution schemes avoid such extra costs yet heavily rely on elaborately designed renoising schedules and fragile hyperparameter configurations, which compromise their deployment stability and limit the final acceleration gains. More importantly, during partial upsampling, existing dynamic resolution frameworks only prioritize edge regions or areas with large channel variance\cite{jeong2025upsample,zheng2026sketch}, thus completely failing to maintain semantic alignment with the input editing guidance.

\section{Method}
\label{sec:method}
\subsection{Preliminaries}
\subsubsection{Diffusion Models}
Diffusion models~\cite{ho2020denoising} define a forward noising process and a learned reverse denoising process. 
The forward process gradually perturbs a clean sample $x_0$ into noise:
\begin{equation}
x_t = \sqrt{\bar{\alpha}_t} x_0 
+ \sqrt{1-\bar{\alpha}_t}\,\epsilon_t,
\quad
\epsilon_t \sim \mathcal{N}(0,\mathbf{I}),
\end{equation}
where $\bar{\alpha}_t=\prod_{s=1}^{t}\alpha_s$. 
The reverse predicts $x_{t-1}$ using a neural network $\epsilon_\theta$:
\begin{equation}
x_{t-1} =
\frac{1}{\sqrt{\alpha_t}}
\left(
x_t - 
\frac{1-\alpha_t}{\sqrt{1-\bar{\alpha}_t}}
\epsilon_\theta(x_t,t)
\right)
+ \sigma_t \epsilon .
\end{equation}
In diffusion Transformers, each step operates over a spatial token grid, leading to quadratic complexity with respect to token count.

\subsubsection{Dynamic Resolution}
Dynamic resolution reduces computation by running early denoising steps on downsampled latents. 
Given $\mathbf{x}_t \in \mathbb{R}^{H\times W\times C}$:
\begin{equation}
\hat{\mathbf{x}}_t =
\mathcal{U}_s
\big(
f(\mathcal{D}_s(\mathbf{x}_t))
\big),
\end{equation}
where $\mathcal{D}_s$ and $\mathcal{U}_s$ denote downsampling and resolution restoration operators. 

For editing, however, the central problem is not merely resolution scaling, but identifying which tokens truly require high-resolution computation.

\subsection{SpecEdit Overview}
\begin{figure}[htbp]
    \centering
    \includegraphics[trim=20 10 10 0, clip,width=1\linewidth]{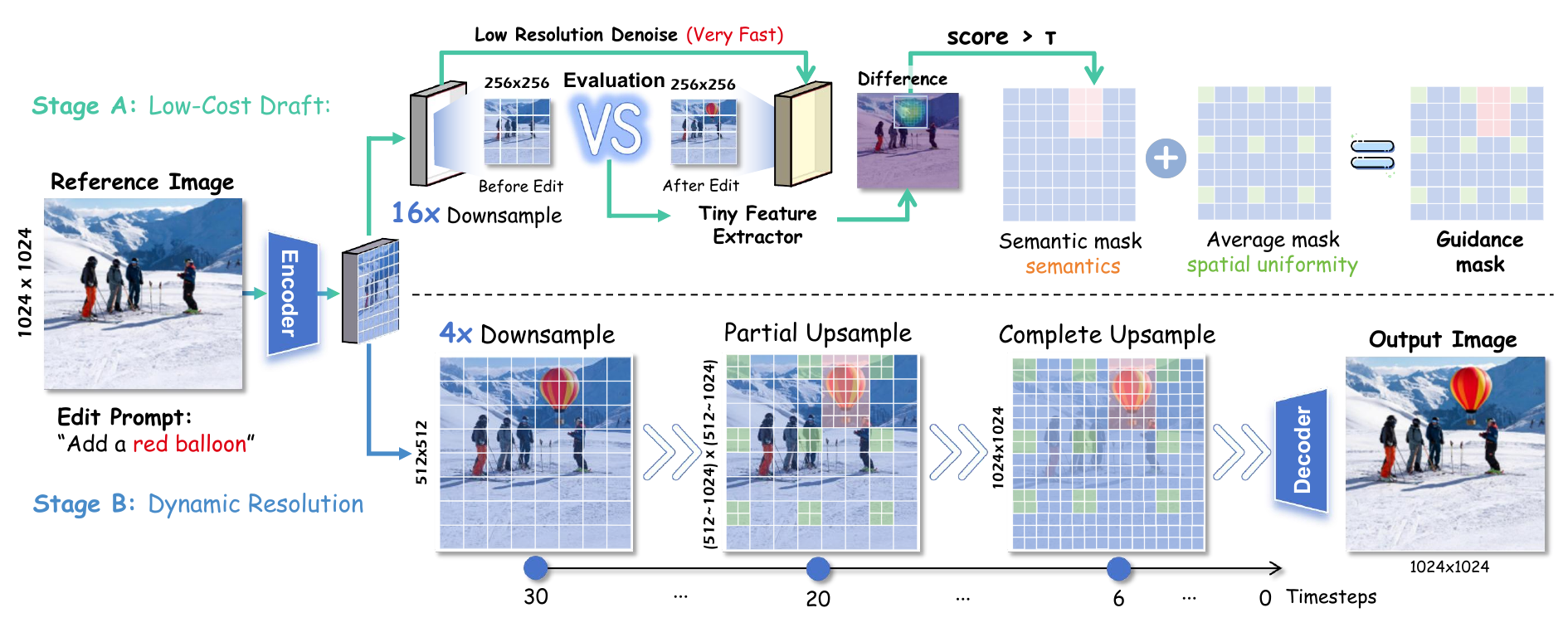}
    \caption{\textbf{Overview of SpecEdit.} A low-resolution draft is produced to estimate semantic discrepancies. The verified tokens are prioritized for high-resolution upsampling during sampling. A sparse set of uniformly sampled tokens is additionally included to maintain global structural stability.}
    \label{fig:method_framework}
\end{figure}

The overall pipeline is illustrated in Fig.~\ref{fig:method_framework}. 
SpecEdit consists of two stages: a preliminary draft stage for semantic verification and a subsequent dynamic-resolution sampling framework for efficient denoising.

\paragraph{Preliminary draft stage.}
Before entering the dynamic-resolution framework, we perform a lightweight draft inference at heavily downsampled ($16\times$) resolution. 
A complete low-resolution denoising trajectory produces a coarse edited result, which is compared with the original input to estimate token-level semantic discrepancies. 
This stage determines the verified tokens requiring high-resolution processing, but does not participate in the subsequent denoising trajectory.

\paragraph{Dynamic-resolution sampling.}
After region verification, we enter the dynamic-resolution framework. 
The latent is first downsampled by a moderate factor ($4\times$) to construct a coarse global representation. 
Verified tokens are selectively expanded to fine resolution for high-resolution Transformer computation, while remaining tokens stay at the coarse resolution. 
In the final reconstruction stage, all tokens are restored to full resolution to produce the output image.

\subsection{Spatial Draft-and-Verify Mechanism}
\subsubsection{Draft Generation}
Before entering the dynamic-resolution denoising framework, 
we introduce a lightweight preliminary stage operating entirely at a heavily downsampled resolution.

Given the input latent $z^{\mathrm{ori}}$, 
we first apply a $16\times$ spatial downsampling:
\begin{equation}
\tilde{z}^{\mathrm{ori}} = \mathcal{D}_{16}(z^{\mathrm{ori}}).
\end{equation}
We then perform a complete diffusion sampling process at this coarse resolution using a fixed number of inference steps:
\begin{equation}
\tilde{z}^{\mathrm{draft}}
=
\mathrm{Denoise}
(
\tilde{z}^{\mathrm{ori}},
\mathbf{c};
\epsilon_\theta
).
\end{equation}
This stage produces a fully denoised low-resolution edited result under standard diffusion inference, but at significantly reduced computational cost. 
The resulting draft is used only for semantic discrepancy estimation.

\subsubsection{Semantic Verification}
To estimate which spatial tokens are likely involved in the edit, 
we compute a token-wise discrepancy map between the draft result and the original input in a perceptual feature space\cite{qin2025spotedit}. 
Specifically, we extract intermediate feature maps $\{\Phi_\ell\}_{\ell=1}^L$ from the decoder and measure a normalized multi-layer feature distance:
\begin{equation}
S(i,j)
=
\frac{1}{L}
\sum_{\ell=1}^{L}
\left\|
\mathrm{Norm}\!\left(\Phi_\ell(\mathcal{U}_s(\tilde{z}^{\mathrm{draft}}))\right)_{i,j}
-
\mathrm{Norm}\!\left(\Phi_\ell(z^{\mathrm{ori}})\right)_{i,j}
\right\|_2^2 ,
\end{equation}
where $\mathrm{Norm}(\cdot)$ denotes channel-wise $\ell_2$ normalization.

Tokens with large discrepancy values are regarded as edit-relevant:
\begin{equation}
\mathcal{T}_{\mathrm{edit}}
=
\{(i,j)\mid S(i,j)>\tau\}.
\end{equation}
This verification step serves as a lightweight semantic estimator for guiding partial upsampling. 
The perceptual discrepancy itself follows standard practice. Our main contribution lies in how the resulting token set is integrated into the dynamic-resolution denoising framework.

\subsection{Resolution Restoration and Selective Computation}
\subsubsection{Token Expansion (Resolution Restoration)}
For tokens selected for high-resolution processing, we restore spatial resolution through a structured token expansion. 
Each coarse token is projected using an orthogonal transformation and combined with a small random perturbation to produce four fine-resolution tokens arranged on a $2\times2$ grid:
\begin{equation}
z^{\mathrm{fine}}_{i,j}
=
\mathrm{Expand}(z^{\mathrm{coarse}}_{i,j}),
\end{equation}
where $\mathrm{Expand}(\cdot)$ denotes a resolution restoration operator that preserves energy under an orthogonal mapping. 
This step merely restores spatial granularity and does not introduce additional learnable parameters in our framework.

\subsubsection{Uniform Coverage}
To stabilize global geometry, we additionally introduce a sparse uniform set:
\begin{equation}
\mathcal{T}_{\mathrm{uniform}}
=
\{(i,j)\mid i\bmod k=0,\ j\bmod k=0\}.
\end{equation}
The final expansion set is:
\begin{equation}
\mathcal{T}_{\mathrm{expand}}
=
\mathcal{T}_{\mathrm{edit}}
\cup
\mathcal{T}_{\mathrm{uniform}}.
\end{equation}

\subsubsection{Selective High-Resolution Transformer Update}
We construct a mixed-resolution latent where tokens in $\mathcal{T}_{\mathrm{expand}}$ are restored to fine resolution, while others remain in draft form. 
Let $\mathcal{M}_t$ denote one Transformer denoising step:
\begin{equation}
z_{t-1}
=
\mathcal{M}_t
(
z^{\mathrm{mixed}}_{t-1},
\mathbf{c},
t;
\mathcal{T}_{\mathrm{expand}}
).
\end{equation}
Only tokens in $\mathcal{T}_{\mathrm{expand}}$ undergo full-resolution attention and feed-forward computation. 
All remaining tokens reuse coarse representations, preserving global context while significantly reducing computation.

SpecEdit differs from prior dynamic-resolution methods in that spatial computation is allocated based on semantic verification rather than geometric heuristics. 
The resolution restoration operator is lightweight and orthogonal to our main contribution. 
The overall framework is fully training-free and compatible with existing temporal acceleration techniques.

\section{Experiment}
\subsection{Experiment Settings}
\textbf{Implementation Details.}
We evaluate SpecEdit on two representative diffusion-based editing models, \textbf{Qwen-Image-Edit} and \textbf{FLUX.1-Kontext-dev}. All experiments are conducted on NVIDIA A100 GPUs.
We compare against diverse acceleration baselines spanning step reduction, attention sparsification\cite{zhang2025spargeattn}, spatial resolution scheduling (e.g., RALU~\cite{jeong2025upsample}, Bottleneck Sampling\cite{tian2025trainingfreediffusionaccelerationbottleneck} and Fresco~\cite{zheng2026sketch}), feature caching and forecasting methods (e.g., ToCa~\cite{zou2024accelerating}, FORA~\cite{selvaraju2024fora}, DuCa~\cite{zou2024DuCa}, Freqca~\cite{liu2025freqca}, TaylorSeer~\cite{liu2025reusing}, FoCa~\cite{zheng2025forecast}), as well as distillation-based variants when applicable. 
To assess compatibility, we further combine SpecEdit with the step-distilled \textbf{FLUX.1-Kontext-Lightning}.

\noindent\textbf{Datasets and Metrics.}
Experiments are conducted on two public editing benchmarks: \textbf{GEdit-Bench}~\cite{liu2025step1x} and \textbf{ImgEdit-Bench}~\cite{ye2025imgedit}, where we focus on the single-turn setting across nine editing categories. 
We evaluate both editing quality and computational efficiency, measuring efficiency by NFE, latency, speedup, and FLOPs, and assessing quality using Structural Consistency (SC), Perceptual Quality (PQ), and Overall Score (OS) on GEdit-Bench, together with category-level and overall performance on ImgEdit-Bench.

\begin{center}  
\setlength{\tabcolsep}{4pt}
\small
\captionof{table}{Quantitative results of image-editing on GEdit-Bench. }
\label{tab:gedit_qwen}
\resizebox{\textwidth}{!}{
  \begin{tabular}{l c c c c c c c c c c c}
    \toprule
    \multirow{2}{*}{\textbf{Method}}
    & \multirow{2}{*}{\textbf{NFE}}
    & \multicolumn{4}{c}{\textbf{Acceleration}}
    & \multicolumn{3}{c}{\textbf{GEdit-CN(FULL)}}
    & \multicolumn{3}{c}{\textbf{GEdit-EN(FULL)}} \\
    \cmidrule(lr){3-6} \cmidrule(lr){7-9} \cmidrule(lr){10-12}
    &
    & \textbf{Latency ($\text{s}$) $\downarrow$}
    & \textbf{Speed $\uparrow$}
    & \textbf{FLOPs ($\text{T}$) $\downarrow$}
    & \textbf{Speed $\uparrow$}
    & \textbf{SC $\uparrow$}
    & \textbf{PQ $\uparrow$}
    & \textbf{OS $\uparrow$}
    & \textbf{SC $\uparrow$}
    & \textbf{PQ $\uparrow$}
    & \textbf{OS $\uparrow$} \\
    \midrule  

    $100\%$ steps
    & 50 & 284.51 & 1.00$\times$ & 28219.71 & 1.00$\times$ & 7.68 & 7.51 & 7.41 & 7.82 & 7.54 & 7.54 \\

    $60\%$ steps
    & 30 & 172.43 & 1.65$\times$ & 16931.83 & 1.67$\times$ & 7.70 & 7.53 & 7.44 & 7.77 & 7.52 & 7.47 \\

    $20\%$ steps
    & 10 & 58.66  & 4.85$\times$ & 5638.18  & 5.00$\times$ & 7.65 & 7.42 & 7.35 & 7.73 & 7.46 & 7.44 \\

    \midrule

    SpargeAttention
    & 50 & 231.30 & 1.23$\times$ & 16846.78 & 1.67$\times$ & 7.87 & 7.57 & 7.56 & 7.81 & 7.53 & 7.50 \\

    ToCa ($\mathcal{N}=6$) 
    & 50 & 172.43 & 1.65$\times$ &  \textbf{7850.13}  &  \textbf{3.59$\times$}         & 7.89   &  7.50  & 7.57   & \textbf{7.89}   & 7.46   &  \textbf{7.54}   \\
    
    Bottleneck Sampling
    & 30 & 109.38 & 2.60$\times$ & 9954.36 & 2.83$\times$ & 7.44 & \textbf{7.59} & 7.28 & 7.62 & 7.40 & 7.24 \\

    RALU
    & 30 & 105.87 & 2.69$\times$ & 9286.52 & 3.04$\times$ & 7.83 & 7.58 & 7.55 & 7.83 & 7.52 & 7.52 \\

    \rowcolor{gray!15}  
    \textbf{SpecEdit}
    & 30 & \textbf{96.23}   & \textbf{2.96$\times$} & 8623.70   & 3.27$\times$ & \textbf{7.91} & 7.53 & \textbf{7.58} & 7.87 & \textbf{7.54} & \textbf{7.54} \\

    \midrule
    Bottleneck Sampling
    & 18 & 85.95     & 3.31$\times$   & 8090.07      & 3.49$\times$        & 7.75          & 7.52   & 7.45      & 7.70   & 7.46   & 7.39   \\

    Duca ($\mathcal{N}=7$)
    & 50 & 69.54  & 4.09$\times$ & 5699.89 & 4.95$\times$ & 7.73 & 7.44 & 7.44 & 7.80 & 7.40 & 7.45 \\
    
    RALU
    & 18 & 66.94 & 4.25$\times$       & 6200.73       & 4.55$\times$          & 7.89   & \textbf{7.56} & \textbf{7.60}   & 7.82   &  \textbf{7.52}   & 7.51  \\
    
    TaylorSeer ($\mathcal{N}=6$)
    & 50 & 65.66  & 4.33$\times$ & 5643.13 & 5.00$\times$ &7.53 & 7.40 & 7.25 & 7.60 & 7.37 & 7.30 \\

    FORA($\mathcal{N}=5$)
    & 50 & 63.15  & 4.51$\times$ & 5643.13   & 5.00$\times$ & 7.60 & 7.31 & 7.25 & 7.62 & 7.34 & 7.28 \\

    \rowcolor{gray!15} 
    \textbf{SpecEdit} ($\mathcal{N}=18$)
    & 18 & \textbf{56.01}     & \textbf{5.08$\times$}    & \textbf{5230.34}    & \textbf{5.40$\times$} & \textbf{7.91} & 7.54 & 7.59 & \textbf{7.92} & \textbf{7.52} & \textbf{7.59} \\

    \midrule
    TaylorSeer ($\mathcal{N}=9$) 
    & 50 & 53.92     & 5.28$\times$    & 4515.74    & 6.25$\times$ & 6.61 & 6.65 & 6.31 & 6.67 & 6.63 & 6.31 \\

    FORA ($\mathcal{N}=7$) 
    & 50 & 52.20  & 5.45$\times$ & 4515.74    & 6.25$\times$ & 7.42 & 7.13 & 7.06 & 7.43 & 7.19 & 7.06 \\

    Freqca ($\mathcal{N}=9$)
    & 50 & 51.09  & 5.57$\times$ & 4514.48    & 6.25$\times$ & 7.62 & 7.18 & 7.27 & 7.66 & 7.12 & 7.21 \\

    \rowcolor{gray!15} 
    \textbf{SpecEdit}
    & 11 & 47.11     & 6.04$\times$    & 3992.31 & 7.07$\times$ & \textbf{7.85} & \textbf{7.51} & \textbf{7.52} & \textbf{7.85} & \textbf{7.51} & \textbf{7.53} \\

    \rowcolor{gray!15} 
    \textbf{SpecEdit}
    & 9  & \textbf{44.27}     & \textbf{6.43$\times$}    & \textbf{2580.52} & \textbf{10.94$\times$} & 7.80 & 7.43 & 7.45 & 7.79 & 7.39 & 7.43 \\

    \bottomrule  
  \end{tabular}
}

{\scriptsize
\begin{itemize}[leftmargin=10pt,topsep=-2pt,label=\textbf{•}]
\item  \textbf{SC}: semantic consistency, \textbf{PQ}: perceptual quality, \textbf{OS}: overall score. 
\end{itemize}
}
\end{center}
\subsection{Results on Qwen-Image-Edit}
\paragraph{Low-to-moderate acceleration.}
In the low-to-moderate regime, SpecEdit consistently achieves the best balance between efficiency and quality.
On \textbf{GEdit-Bench}, at \textbf{30 NFE} it delivers \textbf{2.96$\times$} speedup (96.23s) while achieving the highest overall scores on both CN/EN (OS 7.58/7.54) and strong semantic consistency (SC 7.91/7.87), outperforming \emph{Bottleneck Sampling} (2.60$\times$, OS 7.28/7.24).
On \textbf{ImgEdit-Bench}, SpecEdit matches the full-step overall score (3.88) at 2.96$\times$ speedup, whereas \emph{RALU} at similar acceleration (2.69$\times$) drops to 3.44.
At \textbf{18 NFE}, SpecEdit reaches 5.08$\times$ speedup while maintaining strong performance (OS 7.59/7.59 on GEdit, 3.80 on ImgEdit), showing that semantics-aware token selection preserves fidelity under reduced budgets.
These results indicate that draft-guided semantic verification enables more accurate allocation of high-resolution computation to truly edited regions.
In contrast, heuristic upsampling strategies tend to prioritize geometrically salient areas, which may not align with the actual editing semantics.
Our approach instead identifies semantically relevant tokens through perceptual feature discrepancies, enabling computation to focus on regions that truly participate in the edit.
\begin{center}
  \setlength{\tabcolsep}{3pt}  
  \footnotesize  
  \captionof{table}{Quantitative results of image-editing  on ImgEdit-Bench.}
  \label{tab:imgedit_qwen}
  \resizebox{\linewidth}{!}{
  \begin{tabular}{l c >{\extracolsep{-2pt}}c c<{\extracolsep{0pt}} *{10}{c}} 
    \toprule 
    \multirow{2}{*}{\textbf{Model}} & 
    \multirow{2}{*}{\textbf{NFE}} & 
    \multicolumn{2}{c}{\textbf{Acceleration}} & 
    \multicolumn{10}{c}{\textbf{ImgEdit-Bench}} \\
    \cmidrule(lr){3-4} \cmidrule(lr){5-14} 
    & & \textbf{Latency (s) $\downarrow$} & \textbf{Speedup $\uparrow$} &
    \textbf{Action} & \textbf{Add} & \textbf{Adjust} & \textbf{Background} & 
    \textbf{Compose} & \textbf{Extract} & \textbf{Remove} & \textbf{Replace} & 
    \textbf{Style} & \textbf{Overall $\uparrow$} \\
    \midrule
    \textbf{$100\%$ steps}  & 50 &  284.51 &  1.00$\times$ & 3.97 & 4.72 & 3.96 & 4.19 & 3.57 & 4.25 & 3.97 & 3.14 & 2.92 & 3.88 \\ 
    \textbf{$60\%$ steps}  & 30& 172.43 &1.65$\times$  & 4.01 & 4.63 & 4.15 & 4.07 & 3.57 & 4.34 & 3.93 & 3.08 & 2.94 & 3.88 \\
    \textbf{$20\%$ steps}  & 10 &58.66  & 4.85$\times$ & 3.82 & 4.64 & 3.99 & 3.91 & 3.52 & 4.11 & 3.61 & 2.99 & 2.94 & 3.75 \\
    \midrule
    ToCa (N=6)                & 50 & 172.43 & 1.65$\times$ & 3.93 & 4.56 & \textbf{4.22} & \textbf{4.21} & 3.52 & \textbf{4.32} & 3.84 & 3.04 & 2.96 & 3.88 \\
    Bottleneck                & 30 & 109.38 & 2.60$\times$ & \textbf{3.96} & \textbf{4.68} & 4.16 & 3.95 & \textbf{3.65} & 3.96 & 3.57 & 2.91 & 2.92 & 3.75 \\
    RALU                      & 30 & 105.87 & 2.69$\times$ & 3.87 & 4.40 & 3.55 & 3.77 & 3.13 & 3.72 & 2.93 & 2.92 & 2.63 & 3.44 \\
    \textbf{SpecEdit}                  & 30 & \textbf{96.23} & \textbf{2.96$\times$} & 3.94 & 4.63 & 4.08 & 4.17 & 3.52 & 4.19 & \textbf{4.02} & \textbf{3.10} & \textbf{3.00} & \textbf{3.88} \\
    \midrule
    Bottleneck                & 18 & 85.95 & 3.31$\times$ & 3.65 & 4.13 & 3.15 & 3.88 & 2.93 & 3.49 & 2.63 & 2.67 & 2.75 & 3.25 \\
    FoCa (N=9)                & 50 & 74,67 & 3.81$\times$ & 3.58 & 3.83 & 2.52 & 2.88 & 2.52 & 2.98 & 2.79 & 2.39 & 2.24 & 2.86 \\
    RALU              & 18 & 66.94 & 4.25$\times$ & \textbf{3.97} & 4.52 & 4.01 & 4.03 & \textbf{3.52} & 4.25 & \textbf{3.79} & 2.77 & \textbf{2.94} & 3.76 \\
    \rowcolor{gray!15} 
    \textbf{SpecEdit}                  & 18 & \textbf{56.01} & \textbf{5.08$\times$} & 3.79 & \textbf{4.60} & \textbf{4.08} &\textbf{ 4.06} & 3.50 & \textbf{4.28} & 3.74 & \textbf{3.02} & 2.87 & \textbf{3.80} \\
    \midrule
    DuCa (N=8)                & 50 & 77.73 & 3.66$\times$ & 3.93 & 4.54 & 3.92 & 3.95 & 3.54 & 4.31 & 3.68 & 3.02 & 2.86 & 3.77 \\
    Freqca (N=5)              & 50 & 71.30 & 3.99$\times$ & \textbf{3.96} & \textbf{4.63} & 4.01 & 4.07 & \textbf{3.65} & 4.17 & 3.75 & \textbf{3.21} & \textbf{3.04} & 3.83 \\
    FORA (N=6)                & 50 & 60.40 & 4.71$\times$ & 3.51 & 4.26 & 3.27 & 2.62 & 3.20 & 2.95 & 2.72 & 2.39 & 2.09 & 2.94 \\
    \rowcolor{gray!15} 
    \textbf{SpecEdit}          & 11 & 47.11 & 6.04$\times$ & 3.92 & 4.50 & \textbf{4.01} & \textbf{4.11} & 3.59 & \textbf{4.22} & \textbf{3.98} & 3.02 & 3.00 & \textbf{3.84} \\
    \rowcolor{gray!15} 
    \textbf{SpecEdit}           & 9  & \textbf{44.27} & \textbf{6.43$\times$} & 3.95 & 4.53 & 3.95 & 3.97 & 3.61 & 4.20 & 3.91 & 2.93 & 2.89 & 3.78 \\
    \bottomrule 
  \end{tabular}}
\end{center}

\paragraph{High acceleration.}
SpecEdit remains notably stable.
On \textbf{GEdit-Bench}, at \textbf{11/9 NFE} it achieves 6.04$\times$/6.43$\times$ speedup while maintaining competitive quality (OS 7.52/7.45 on CN and 7.53/7.43 on EN).
In contrast, \emph{TaylorSeer} (N=9) attains comparable speedup (5.28$\times$) but degrades substantially (OS 6.31/6.31).
On \textbf{ImgEdit-Bench}, SpecEdit maintains 3.84 (11 NFE) and 3.78 (9 NFE) at over 6$\times$ acceleration.
These results indicate that draft-guided semantic verification enables robust structural preservation even in the high-acceleration regime.
\begin{center}  
\setlength{\tabcolsep}{4pt}
\small
\captionof{table}{Compatibility with Acceleration Methods on GEdit-Bench. }  
\label{tab:qwen_specedit_cache}
\resizebox{\textwidth}{!}{
  \begin{tabular}{l c c c c c c c c c c c}
    \toprule
    \multirow{2}{*}{\textbf{Method}}
    & \multirow{2}{*}{\textbf{NFE}}
    & \multicolumn{4}{c}{\textbf{Acceleration}}
    & \multicolumn{3}{c}{\textbf{GEdit-CN(FULL)}}
    & \multicolumn{3}{c}{\textbf{GEdit-EN(FULL)}} \\
    \cmidrule(lr){3-6} \cmidrule(lr){7-9} \cmidrule(lr){10-12}
    &
    & \textbf{Latency ($\text{s}$) $\downarrow$}
    & \textbf{Speed $\uparrow$}
    & \textbf{FLOPs ($\text{T}$) $\downarrow$}
    & \textbf{Speed $\uparrow$}
    & \textbf{SC $\uparrow$}
    & \textbf{PQ $\uparrow$}
    & \textbf{OS $\uparrow$}
    & \textbf{SC $\uparrow$}
    & \textbf{PQ $\uparrow$}
    & \textbf{OS $\uparrow$} \\
    \midrule  

    $100\%$ steps
    & 50 & 284.51 & 1.00$\times$ & 28219.71 & 1.00$\times$ & 7.68 & \textbf{7.51} & 7.41 & 7.82 & \textbf{7.54} & 7.54 \\

    \midrule

    \rowcolor{gray!15}  
    \textbf{SpecEdit+FoCa}
    & 18 &\textbf{47.45}    & \textbf{6.00$\times$} &\textbf{4453.91}    &\textbf{6.34$\times$} & \textbf{7.86} & \textbf{7.51} & \textbf{7.56} &\textbf{7.87} &7.53  & \textbf{7.57} \\

    \bottomrule  
  \end{tabular}
}
  {\scriptsize
\begin{itemize}[leftmargin=10pt,topsep=-2pt,label=\textbf{•}]
\item  \textbf{SC}: semantic consistency, \textbf{PQ}: perceptual quality, \textbf{OS}: overall score. 
\end{itemize}
}
\end{center} 
\begin{center}
  \setlength{\tabcolsep}{3pt}  
  \footnotesize  
  \captionof{table}{ Compatibility with Acceleration Methods  on ImgEdit-Bench.}
  \label{tab:qwen_specedit_cache_imgedit}
  \resizebox{\linewidth}{!}{
  \begin{tabular}{l c >{\extracolsep{-2pt}}c c<{\extracolsep{0pt}} *{10}{c}} 
    \toprule 
    \multirow{2}{*}{\textbf{Model}} & 
    \multirow{2}{*}{\textbf{NFE}} & 
    \multicolumn{2}{c}{\textbf{Acceleration}} & 
    \multicolumn{10}{c}{\textbf{ImgEdit-Bench}} \\
    \cmidrule(lr){3-4} \cmidrule(lr){5-14} 
    & & \textbf{Latency (s) $\downarrow$} & \textbf{Speedup $\uparrow$} &
    \textbf{Action} & \textbf{Add} & \textbf{Adjust} & \textbf{Background} & 
    \textbf{Compose} & \textbf{Extract} & \textbf{Remove} & \textbf{Replace} & 
    \textbf{Style} & \textbf{Overall$\uparrow$} \\
    \midrule
    $100\%$ steps  & 50 &  284.51 &  1.00$\times$ & \textbf{3.97} & \textbf{4.72} & 3.96 & \textbf{4.19} & \textbf{3.57} & \textbf{4.25} & \textbf{3.97} & \textbf{3.14} & \textbf{2.92} & \textbf{3.88} \\ 

    \midrule
    \rowcolor{gray!15}  
   \textbf{ SpecEdit + FoCa} & 18 & \textbf{47.45}  & \textbf{6.00 $\times$} & 3.88 & 4.50 & \textbf{4.02} & 3.96 & 3.54 &  4.24 & 3.90 & 2.87 & 2.87 & 3.75 \\

    \bottomrule 
  \end{tabular}}
\end{center}

\paragraph{Compatibility with complementary acceleration methods.}
Table~\ref{tab:qwen_specedit_cache_imgedit} and Table~\ref{tab:qwen_specedit_cache} evaluate the compatibility of SpecEdit with feature caching (FoCa).
On \textbf{ImgEdit-Bench}, combining SpecEdit with FoCa achieves \textbf{6.00$\times$} latency speedup (47.45s) while maintaining competitive overall performance (3.75 vs. 3.88).
On \textbf{GEdit-Bench}, SpecEdit+FoCa further delivers \textbf{6.00$\times$} latency and \textbf{6.34$\times$} FLOPs reduction while sustaining strong semantic consistency and perceptual quality (OS 7.56/7.57 on CN/EN), close to the full-step baseline.

\begin{center}
  \setlength{\tabcolsep}{4pt}  
  \footnotesize  
  \captionof{table}{Quantitative results of image-editing on GEdit-Bench. }
  \label{tab:gedit_flux}
  \resizebox{\textwidth}{!}{
    \begin{tabular}{l c c c c c c c c}
      \toprule  
      \multirow{2}{*}{\textbf{Method}} & 
      \multirow{2}{*}{\textbf{NFE}} & 
      \multicolumn{4}{c}{\textbf{Acceleration}} & 
      \multicolumn{3}{c}{\textbf{GEdit-EN(FULL)}} \\
      \cmidrule(lr){3-6} \cmidrule(lr){7-9}  
      & & \textbf{Latency ($\text{s}$) $\downarrow$} & \textbf{Speed $\uparrow$} &
      \textbf{FLOPs ($\text{T}$) $\downarrow$} & \textbf{Speed $\uparrow$} &
      \textbf{SC $\uparrow$} & \textbf{PQ $\uparrow$} & \textbf{OS $\uparrow$} \\
      \midrule  

      $100\%$ steps & 50 & 50.20 & 1.00$\times$ & 8299.54 & 1.00$\times$ & 6.80  & 7.26  & 6.51  \\
      $60\%$ steps  & 30 & 32.23 & 1.56$\times$ & 4979.72 & 1.67$\times$ & 6.54 & 7.28 & 6.25 \\
      $20\%$ steps  & 10 & 10.47 & 4.79$\times$ & 1659.91 & 5.00$\times$ & 6.60 & 7.18 & 6.28 \\

      \midrule
      SpargeAttention         & 50 & 46.05 & 1.09$\times$ & 5603.68 & 1.48$\times$ & 6.46  & \textbf{7.25}  & 6.21  \\
      RALU     & 30 & 23.34             & 2.15$\times$             & 2730.53             & 3.04$\times$             & 7.06  & 7.20  & \textbf{6.70}  \\  
     Fresco& 30 &20.48  & 2.45$\times$ & \textbf{2361.22} & \textbf{3.51$\times$}  & 7.02  & 7.12  & 6.65  \\  
      Bottleneck Sampling & 30 & 18.25 & 2.75$\times$ & 2727.10 & 3.04$\times$ & 6.46 & 6.63 & 6.08 \\  
      \rowcolor{gray!15}  
      \textbf{SpecEdit}                 & 30 & \textbf{15.93}             & \textbf{3.15$\times$}             & 2508.60             & 3.31$\times$             & \textbf{7.08}  & 7.08 & 6.65 \\

      \midrule
      ToCa ($\mathcal{N}=8$) & 50 & 29.56 & 1.70$\times$ & 1841.35 & 4.51$\times$ & 6.43  & 7.25  & 6.12  \\
      Fresco & 18 &17.73   & 2.83$\times$ & 1878.83 & 4.42$\times$ & 7.01  & 7.15  & 6.66  \\  
      RALU     & 18 &  15.30            &   3.28$\times$                    & 2123.74                & 3.91$\times$                      & 7.05    & 7.16    & 6.68    \\  
      Bottleneck Sampling  & 18 & 15.21 & 3.30$\times$ & 2274.58 & 3.65$\times$ & 6.86 & 6.77 & 6.43 \\  
      
      TaylorSeer ($\mathcal{N}=6$) & 50 & 13.95 & 3.60$\times$ & 1660.95 & 5.00$\times$ & 6.47  & \textbf{7.29}  & 6.17  \\  
      FoCa ($\mathcal{N}=6$) & 50 & 13.86 & 3.62$\times$ & 1660.01 & 5.00$\times$ & 6.62  & 7.27  & 6.27  \\  


      \rowcolor{gray!15}  
      \textbf{SpecEdit}                 & 18 & \textbf{12.24}    & \textbf{4.10$\times$}    & \textbf{1541.50 }    & \textbf{5.38$\times$}             & \textbf{7.12} & 6.98 & \textbf{6.69} \\

      \midrule
      ToCa ($\mathcal{N}=12$) & 50 & 20.72 & 2.42$\times$ & 1359.61 & 6.10$\times$ & 6.39 & 6.91 & 6.04 \\
      TaylorSeer ($\mathcal{N}=9$) & 50 & 12.05 & 4.17$\times$ & 1329.02 & 6.24$\times$ & 6.40 & 6.99 & 6.07 \\  
      DuCa ($\mathcal{N}=12$) & 50 & 10.39 & 4.83$\times$ & 1376.55 & 6.03$\times$ & 6.59 & 7.01 & 6.17 \\  
      \rowcolor{gray!15}  
      \textbf{SpecEdit}                 & 11 & 7.84     & 6.40$\times$    & 1199.78 & 6.92$\times$ & \textbf{7.13} & 6.97 & 6.68 \\

      \rowcolor{gray!15}  
      \textbf{SpecEdit}                 & 9  & \textbf{7.12}     & \textbf{7.05$\times$}    & \textbf{1125.74} & \textbf{7.37$\times$} & 7.01 & \textbf{7.07} & \textbf{6.75} \\

      \bottomrule  
    \end{tabular}
  }
  {\scriptsize
\begin{itemize}[leftmargin=10pt,topsep=-2pt,label=\textbf{•}]
\item  \textbf{SC}: semantic consistency, \textbf{PQ}: perceptual quality, \textbf{OS}: overall score. 
\end{itemize}
}
\end{center}
\subsection{Results on FLUX.1-Kontext-dev}
In the low to moderate regime (30/18 NFE), SpecEdit consistently achieves a strong balance between efficiency and quality.
On \textbf{GEdit-Bench}, at 30 NFE it delivers \textbf{3.15$\times$} speedup with the highest SC (7.08) and competitive OS (6.65), outperforming \emph{Bottleneck Sampling} (2.75$\times$, OS 6.08) and matching or surpassing \emph{Fresco} at higher acceleration.

At 18 NFE, SpecEdit further reaches \textbf{4.10$\times$} speedup with the best OS (6.69), surpassing both RALU and Bottleneck.
On \textbf{ImgEdit-Bench}, SpecEdit achieves 3.15$\times$ and 4.10$\times$ speedup at 30/18 NFE while maintaining the strongest overall performance (3.64/3.68), indicating more stable semantic preservation than heuristic spatial or temporal baselines.

This improvement suggests that draft-guided semantic verification provides a more reliable criterion for token selection, enabling computation to focus on truly edited regions while preserving global structure and maintaining consistent editing quality under different sampling budgets.
Notably, these improvements are reflected across both semantic consistency and overall performance, indicating that the proposed mechanism achieves a favorable balance between efficiency and fidelity.
By reducing redundant high-resolution computation in semantically irrelevant regions, SpecEdit improves the stability of the denoising process and maintains consistent editing behavior across datasets and acceleration levels.

\begin{center}
  \setlength{\tabcolsep}{2pt}  
  \footnotesize  
  \captionof{table}{Quantitative results of image-editing on ImgEdit-Bench.}
  \label{tab:imgedit_flux}  
  \resizebox{\linewidth}{!}{
  \begin{tabular}{l l >{\extracolsep{-3pt}}c c<{\extracolsep{0pt}} *{10}{c}} 
    \toprule  
    \multirow{2}{*}{\textbf{Model}} & 
    \multirow{2}{*}{\textbf{NFE}} & 
    \multicolumn{2}{c}{\textbf{Acceleration}} & 
    \multicolumn{10}{c}{\textbf{ImgEdit-Bench}} \\
    \cmidrule(lr){3-4} \cmidrule(lr){5-14}  
    & & \textbf{Latency ($\text{s}$) $\downarrow$} & \textbf{Speedup $\uparrow$} & 
    \textbf{Action} & \textbf{Add} & \textbf{Adjust} & \textbf{Background} & 
    \textbf{Compose} & \textbf{Extract} & \textbf{Remove} & \textbf{Replace} & 
    \textbf{Style} & \textbf{Overall $\uparrow$} \\
    \midrule  
    \textbf{$100\%$ steps}        & 50 & 50.20 & 1.00$\times$ & 3.89 & 4.41 & 3.60 & 4.05 & 3.24 & 4.00 & 3.47 & 3.20 & 2.84 & 3.66 \\ 
    \textbf{$60\%$ steps}         & 30 & 32.23 & 1.56$\times$ & 3.87 & 4.41 & 3.45 & 4.01 & 3.34 & 3.99 & 3.31 & 3.14 & 2.85 & 3.60 \\ 
    \textbf{$20\%$ steps}        & 10 & 10.47 & 4.79$\times$ & 3.82 & 4.53 & 3.41 & 3.95 & 3.30 & 3.87 & 3.29 & 3.22 & 2.70 & 3.58 \\
    \midrule
    SpargeAttention        &  50  &48.26  &1.04$\times$   & 3.88 & 4.48 & 3.59 & 3.82 & 3.26 & 3.86 & 2.92 & 3.08 & 2.88 & 3.54 \\ 
    RALU  & 30 &23.34  & 2.15$\times$  & 3.86 & \textbf{4.61} & \textbf{3.63} & 4.07 & \textbf{3.28} & \textbf{3.96} & \textbf{3.43} & 3.08 & \textbf{2.91}&\textbf{3.64}  \\
    Bottleneck Sampling  & 30 & 18.25 & 2.75$\times$  & 3.70 & 4.11 & 3.36 & 3.86 & 2.89 & 3.41 & 2.66 & 2.76 & 2.82 & 3.30 \\  
     
    \rowcolor{gray!15} 
    \textbf{SpecEdit}  & 30 & \textbf{15.93} &\textbf{3.15$\times$}  & \textbf{3.92} & 4.51 & 3.52 & \textbf{4.16} & 3.15 & 3.89 & 3.35 & \textbf{3.33} & 2.72& \textbf{3.64}\\
    \midrule
    Fresco          & 18 &17.73 & 2.83$\times$ & 3.82  & 4.53 & 3.44 & \textbf{4.16} & 3.33 & 3.92 & \textbf{3.46} & 3.23 & 2.82 & 3.63  \\
    Bottleneck Sampling  & 18 & 15.21 & 3.30$\times$  & 3.56 & 4.21 & 3.09 & 3.85 & 2.92 & 3.49 & 2.69 & 2.77 &\textbf{2.87} &3.27 \\
    TaylorSeer ($N=6$)& 50& 13.21 & 3.80$\times$  & 3.86 & \textbf{4.62} & 3.55 & 3.63 & 3.24 & 3.76 & 3.07 & 2.71 & 2.52 & 3.44 \\  
    FORA ($N=6$)         & 50 & 12.26 & 4.09$\times$  & 3.86 & 4.38 & 3.50 & 3.84 & \textbf{3.35} & 3.87 & 3.03 & 3.11 & 2.80 & 3.53 \\
    \rowcolor{gray!15} 
    \textbf{SpecEdit}             & 18 & \textbf{12.24} & \textbf{4.10$\times$} & \textbf{3.92} & 4.53 & \textbf{3.59} & 4.15 & 3.24 & \textbf{3.95} & 3.45 & \textbf{3.36} & 2.75 & \textbf{3.68} \\
    \midrule
    TaylorSeer ($N=9$)   & 50 & 11.43 & 4.39$\times$  & 3.60 & 4.17 & 3.11 & 3.72 & 2.93 & 3.41 & 2.78 & 3.03 & \textbf{2.74} & 3.29 \\  
    FORA ($N=9$)         & 50 & 8.61  & 5.83$\times$  & 3.86 & 4.44 & 3.31 & 3.77 & 2.50 & 3.71 & 3.00 & 3.04 & 2.63 & 3.42 \\  
    RALU                & 18 & 9.07  & 5.53$\times$  & 3.87 & 4.40 & 3.55 & 3.77 & 3.13 & 3.72 & 2.93 & 2.92 & 2.63 & 3.44 \\  
    \rowcolor{gray!15} 
    \textbf{SpecEdit}             & 11 & 7.84  & 6.40$\times$ & \textbf{3.88} & \textbf{4.54} & \textbf{3.60} & \textbf{4.06} & \textbf{3.22} & \textbf{3.76} & \textbf{3.41} & \textbf{3.27} & 2.66 & \textbf{3.62} \\
    \rowcolor{gray!15} 
    \textbf{SpecEdit}             & 9  & \textbf{7.12}  & \textbf{7.05$\times$} & 3.82 & 4.42 & 3.47 & 3.94 & 2.76 & 3.71 & 3.28 & 3.09 & 2.54 & 3.49 \\
    \bottomrule  
  \end{tabular}}
\end{center}
\begin{figure}[htbp]
    \centering
    \includegraphics[width=\linewidth]{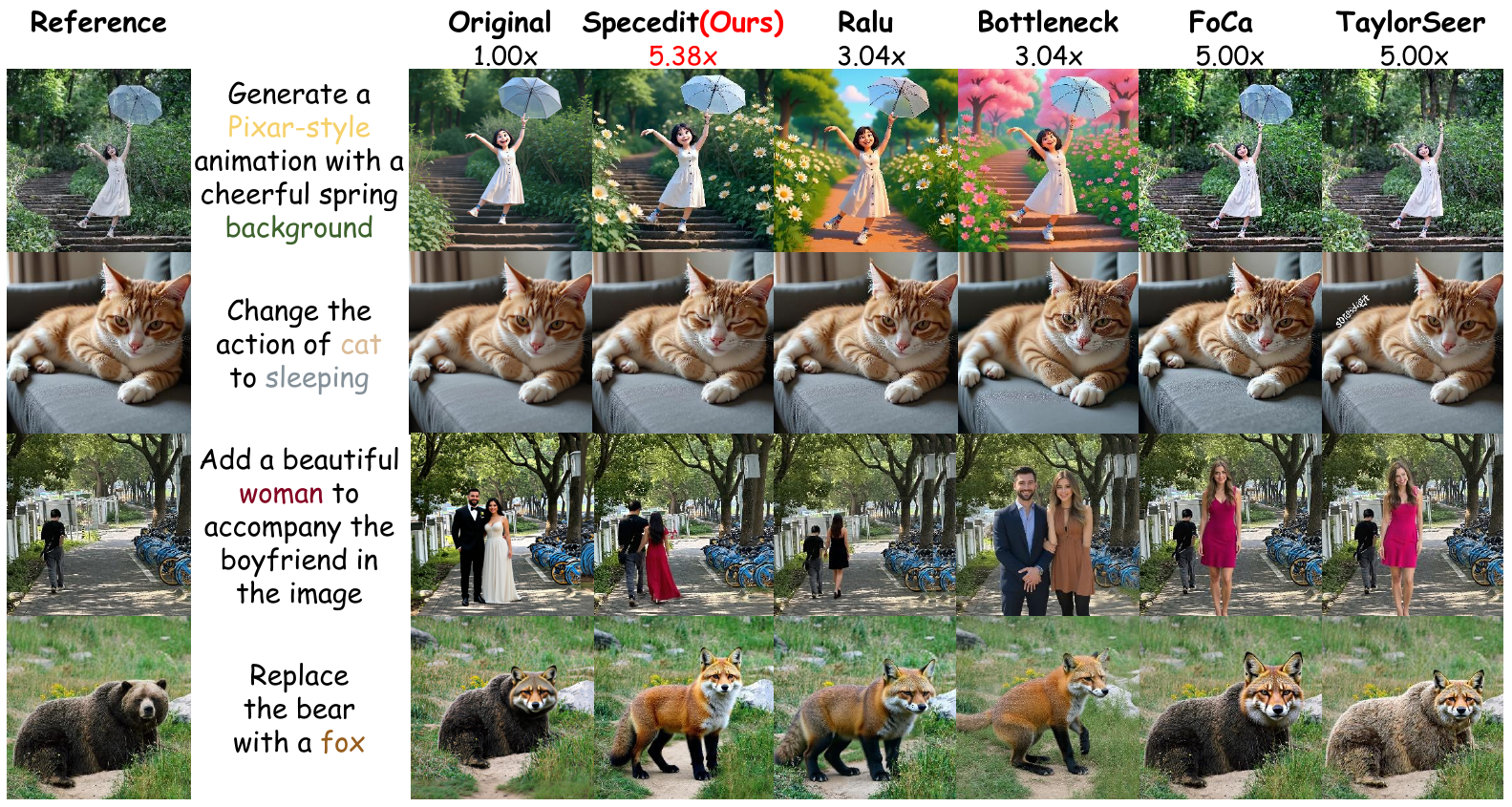}
    \caption{\textbf{Qualitative comparison on GEdit-Bench.} SpecEdit achieves superior semantic preservation and visual quality at a higher acceleration ratio (\textbf{5.38$\times$}) compared to state-of-the-art baselines.}
    \label{fig:specedit_gedit_compare}
\end{figure}

Under aggressive computation budgets (11/9 NFE), \textbf{SpecEdit} remains stable and reliable without noticeable structural degradation.
On \textbf{GEdit-Bench}, it achieves \textbf{6.40$\times$}/\textbf{7.05$\times$} acceleration while maintaining strong perceptual quality (OS 6.68/6.75), substantially outperforming \emph{TaylorSeer} (4.17$\times$, OS 6.07) under comparable settings.
On \textbf{ImgEdit-Bench}, SpecEdit preserves competitive overall performance (3.62/3.49) while achieving more than $6\times$ speedup, indicating a favorable balance between robustness and efficiency.

These results suggest that the draft-guided semantic verification mechanism effectively stabilizes the denoising trajectory, preserving structural consistency and semantic alignment even under high acceleration. Specifically, by first generating a low-resolution draft to accurately identify edit-relevant spatial tokens through perceptual feature discrepancy estimation, the mechanism ensures that high-resolution computational resources are strategically allocated only to the regions that truly require semantic modification. As a result, the edited output maintains robust structural consistency with the original input in unmodified regions while achieving precise semantic alignment with the editing instructions in modified areas, striking a critical balance between efficiency and fidelity that prior heuristic-driven dynamic-resolution approaches fail to achieve.

\begin{center}
  \setlength\tabcolsep{4pt}  
  \small  
  \captionof{table}{ Compatibility with Other Acceleration Methods on GEdit-Bench. }
  \label{table:distill}  
  \resizebox{\textwidth}{!}{
    \begin{tabular}{l c c c c c c c c}
      \toprule  
      \multirow{2}{*}{\textbf{Method}} & \multirow{2}{*}{\textbf{NFE}} 
      & \multicolumn{4}{c}{\textbf{Acceleration}}
      & \multicolumn{3}{c}{\textbf{GEdit-EN(FULL)}} \\
      \cmidrule(lr){3-6} \cmidrule(lr){7-9}  
      & & \textbf{Latency ($\text{s}$) $\downarrow$} & \textbf{Speed $\uparrow$} 
      & \textbf{FLOPs ($\text{T}$) $\downarrow$}  & \textbf{Speed $\uparrow$} 
      & \textbf{SC $\uparrow$} & \textbf{PQ $\uparrow$} & \textbf{OS $\uparrow$} \\
      \midrule  
      FLUX.1-Kontext-dev & 50 & 50.20 & 1.00$\times$ & 8299.54 & 1.00$\times$ & 6.80 & 7.26 & 6.51 \\
      \midrule
      TaylorSeer ($\mathcal{N}=6$) & 50 & 13.95 & 3.60$\times$ & 1660.95 & 5.00$\times$ & 6.47  & \textbf{7.29}  & 6.17  \\  

      \rowcolor{gray!15} 
      \textbf{SpecEdit}+TaylorSeer  & \textbf{30} & \textbf{11.43}  & \textbf{4.39$\times$} & \textbf{1238.73}  & \textbf{6.70$\times$} & \textbf{7.04} & 6.81 & \textbf{6.56} \\
      \midrule

      FLUX.1-Kontext-Lightning  & 8  & 8.24  & 6.09$\times$ & 1327.92 & 6.25$\times$ & 6.91 & \textbf{7.23} & 6.61 \\  
      FLUX.1-Kontext-Lightning  & 6  & 6.36  & 7.89$\times$ & 995.94 & 8.33$\times$ & 6.96 & 7.19 & \textbf{6.63} \\  

      \rowcolor{gray!15} 
      \textbf{SpecEdit}+Step Distillation  & 8  & 6.97 & 7.20$\times$ & 856.98  & 9.68$\times$ & \textbf{6.99} & 7.02 & 6.58 \\

      \rowcolor{gray!15} 
      \textbf{SpecEdit}+Step Distillation  & \textbf{6}  & \textbf{5.48} & \textbf{9.16$\times$} & \textbf{636.65} & \textbf{13.03$\times$} & 6.93 & 7.12 & 6.57  \\

      \bottomrule  
    \end{tabular}
  }
  {\scriptsize
\begin{itemize}[leftmargin=10pt,topsep=-2pt,label=\textbf{•}]
\item  \textbf{SC}: semantic consistency, \textbf{PQ}: perceptual quality, \textbf{OS}: overall score. 
\end{itemize}
}

\end{center}
\begin{center}
  \setlength{\tabcolsep}{2pt}  
  \footnotesize  
  \captionof{table}{Compatibility with Other Acceleration Methods on ImgEdit-Bench. }
  \label{tab:flux_specedit_cache_imgedit}  
  \resizebox{\linewidth}{!}{
  \begin{tabular}{l l >{\extracolsep{-3pt}}c c<{\extracolsep{0pt}} *{10}{c}} 
    \toprule  
    \multirow{2}{*}{\textbf{Model}} & 
    \multirow{2}{*}{\textbf{NFE}} & 
    \multicolumn{2}{c}{\textbf{Acceleration}} & 
    \multicolumn{10}{c}{\textbf{ImgEdit-Bench}} \\
    \cmidrule(lr){3-4} \cmidrule(lr){5-14}  
    & & \textbf{Latency ($\text{s}$) $\downarrow$} & \textbf{Speedup $\uparrow$} & 
    \textbf{Action} & \textbf{Add} & \textbf{Adjust} & \textbf{Background} & 
    \textbf{Compose} & \textbf{Extract} & \textbf{Remove} & \textbf{Replace} & 
    \textbf{Style} & \textbf{Overall$\uparrow$} \\
    \midrule  
    FLUX.1-Kontext-dev         & 50 & 50.20 & 1.00$\times$ & 3.89 & 4.41 & \textbf{3.60} & 4.05 & 3.24 & \textbf{4.00} & \textbf{3.47} & 3.20 & \textbf{2.84} & \textbf{3.66} \\ 
    
    \midrule  
    \rowcolor{gray!15}  
   \textbf{ Specedit+Taylorseer }        & 30 &\textbf{11.43}  & \textbf{4.39} $\times$ & \textbf{3.92} & \textbf{4.47} & 3.46 & \textbf{4.13} & \textbf{3.35} & 3.88 & 3.41 & \textbf{3.36} & 2.72 & 3.65 \\ 
    \bottomrule  
  \end{tabular}}
\vspace{-4pt}
\end{center}

\begin{figure}[h]
    \centering
    \includegraphics[width=\linewidth]{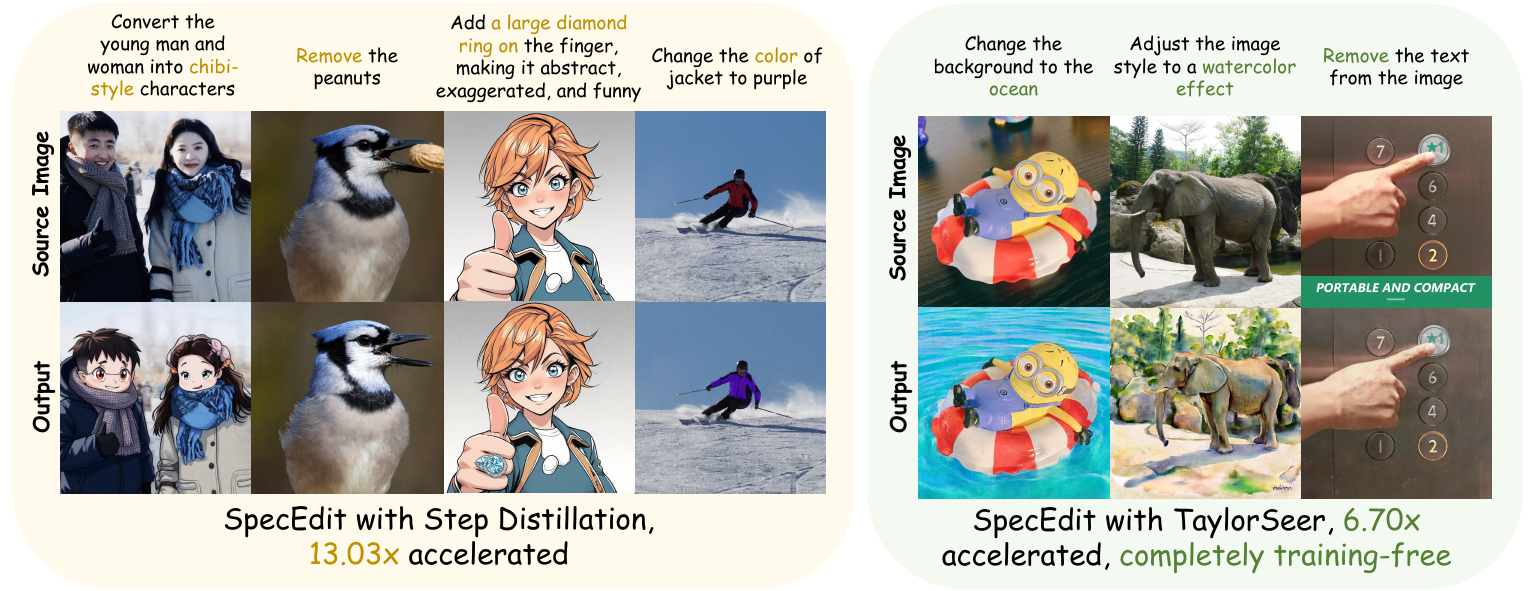}
    \caption{\textbf{Compatibility with other methods.} SpecEdit achieves additive speedups of up to \textbf{13.03$\times$} and \textbf{6.70$\times$} when combined with distillation and caching, respectively.}
    \label{fig:compatibility}
\end{figure}

Table~\ref{table:distill} and Table~\ref{tab:flux_specedit_cache_imgedit} demonstrate that SpecEdit is orthogonal to both temporal caching and step distillation.
On \textbf{GEdit-Bench}, combining SpecEdit with \emph{TaylorSeer} improves latency from 3.60$\times$ to \textbf{4.39$\times$} speedup while increasing SC from 6.47 to \textbf{7.04} and raising OS from 6.17 to \textbf{6.56}, indicating that semantics-aware spatial allocation complements temporal feature reuse.
When integrated with step distillation (FLUX.1-Kontext-Lightning), SpecEdit further achieves up to \textbf{9.16$\times$} latency and \textbf{13.03$\times$} FLOPs reduction, while maintaining competitive quality (OS 6.57 at 6 NFE), close to or exceeding the distilled baseline.
Consistent gains are observed on \textbf{ImgEdit-Bench}, where SpecEdit+TaylorSeer reaches \textbf{4.39$\times$} speedup with stable overall performance (3.65).
These results confirm that draft-guided semantic verification provides additive acceleration without compromising structural consistency or semantic fidelity.

\begin{center}
  \setlength\tabcolsep{4pt}  
  \small  
\captionof{table}{Ablation Study on FLUX.1-Kontext-dev with GEdit-Bench}
  \label{table:upsample_different_method}  
  \resizebox{\textwidth}{!}{
    \begin{tabular}{l c c c c c c c c}
      \toprule  
      \multirow{2}{*}{\textbf{Method}} & \multirow{2}{*}{\textbf{NFE}} 
      & \multicolumn{4}{c}{\textbf{Acceleration}}
      & \multicolumn{3}{c}{\textbf{GEdit-EN(FULL)}} \\
      \cmidrule(lr){3-6} \cmidrule(lr){7-9}  
      & & \textbf{Latency ($\text{s}$) $\downarrow$} & \textbf{Speed $\uparrow$} 
      & \textbf{FLOPs ($\text{T}$) $\downarrow$}  & \textbf{Speed $\uparrow$} 
      & \textbf{SC $\uparrow$} & \textbf{PQ $\uparrow$} & \textbf{OS $\uparrow$} \\
      \midrule  
        FLUX.1-Kontext-dev & 50 & 50.20 & 1.00$\times$ & 8299.54 & 1.00$\times$ & 6.80 & 7.26 & 6.51 \\
      \midrule  
      
      Only semantic Region Upsampling & 18 & 12.15 & 4.13$\times$ & 1518.58 & 5.47$\times$ & 7.10 & 6.80 & 6.60 \\
      Average Upampling & 18 & \textbf{11.59} & \textbf{4.33$\times$} & \textbf{1383.57} & \textbf{5.99$\times$} & 6.94 & \textbf{7.08} & 6.56 \\
        \rowcolor{gray!15}  
      \textbf{SpecEdit} & 18& 12.24    & 4.10$\times$    & 1541.50    & 5.38$\times$ & \textbf{7.12} & 6.98 & \textbf{6.69} \\
      \midrule  
      
      SpecEdit(Average k=2) & 18 & 12.50 & 4.02$\times$ & 1573.77 & 5.27$\times$ & 7.02 & \textbf{7.12} & 6.63 \\
        SpecEdit(Average k=4) & 18 & \textbf{11.80} & \textbf{4.25$\times$} & \textbf{1532.55} & \textbf{5.42$\times$} & 7.12 & 6.95 & 6.65 \\
      \rowcolor{gray!15}  
      \textbf{SpecEdit(Average k=3)} & 18 & 12.24    & 4.10$\times$    & 1541.50    & 5.38$\times$ & \textbf{7.12} & 6.98 & \textbf{6.69} \\
      \midrule  
      Specedit(w/o draft) & 18 & \textbf{12.10}    & \textbf{4.15$\times$}    & \textbf{1449.01}   & \textbf{5.73$\times$} & - & - & - \\
      \rowcolor{gray!15}  
      \textbf{Specedit(w/ draft)} & 18 & 12.24    & 4.10$\times$    & 1541.50    & 5.38$\times$ & \textbf{7.12} & \textbf{6.98} & \textbf{6.69} \\
      \bottomrule  
    \end{tabular}
  }
  {\scriptsize
\begin{itemize}[leftmargin=10pt,topsep=-2pt,label=\textbf{•}]
\item  \textbf{SC}: semantic consistency, \textbf{PQ}: perceptual quality, \textbf{OS}: overall score. 
\end{itemize}
}
\end{center}
\section{Ablation study}
\paragraph{Effect of draft-based verification and upsampling strategies.}
Table~\ref{table:upsample_different_method} compares token upsampling strategies under the same budget (18 NFE) on FLUX.1-Kontext-dev (GEdit-Bench).
Semantic-only upsampling improves semantic consistency (SC 7.10) but reduces perceptual quality (PQ 6.80), while uniform upsampling achieves the highest efficiency (4.33$\times$ latency, 5.99$\times$ FLOPs reduction) and higher PQ (7.08) at the cost of weaker semantic alignment (SC 6.94).
These results reveal a clear trade-off between semantic fidelity and perceptual quality when relying on a single token selection strategy.

By combining semantic verification with sparse uniform coverage, \textbf{SpecEdit} achieves the best overall performance (OS 6.69) and the highest semantic consistency (SC 7.12) while maintaining competitive acceleration (4.10$\times$).
Performance remains stable across different strides $k$ (best at $k=3$), and removing the draft stage leads to unstable token selection, confirming that the ultra-low-resolution draft provides an effective and low-cost semantic prior.

\section{Conclusion}
We presented \textbf{SpecEdit}, a training-free dynamic-resolution framework tailored to diffusion-based image editing. 
SpecEdit follows a draft-and-verify paradigm: a low-resolution draft first approximates the semantic evolution of the edit, and token-level perceptual discrepancies are then used to identify tokens that warrant high-resolution Transformer computation, with a sparse uniform coverage set introduced to stabilize global structure. 
Across Qwen-Image-Edit and FLUX.1-Kontext-dev on GEdit-Bench and ImgEdit-Bench, SpecEdit consistently improves the balance between efficiency and quality, achieving up to \(10\times\) and \(7\times\) acceleration while maintaining strong semantic consistency and perceptual quality. 
The framework is fully training-free and compatible with complementary acceleration techniques such as caching and temporal reduction, providing additional gains when combined. 
We hope this work encourages further research on semantics-aware spatial computation for efficient and high-fidelity image editing.

\bibliographystyle{unsrt}  
\bibliography{main}

\clearpage
\section*{Supplementary Material}
\section{Additional Ablation Studies}

\begin{center}
  \setlength\tabcolsep{4pt}
  \small

  \captionof{table}{Ablation Study on FLUX.1-Kontext-dev with GEdit-Bench}
  \label{table:ablation_flux_kontext_gedit_bench}

  \resizebox{\textwidth}{!}{
    \begin{tabular}{l c c c c c c c c}
      \toprule
      \multirow{2}{*}{\textbf{Method}} 
      & \multirow{2}{*}{\textbf{NFE}} 
      & \multicolumn{4}{c}{\textbf{Acceleration}}
      & \multicolumn{3}{c}{\textbf{GEdit-EN(FULL)}} \\
      \cmidrule(lr){3-6} \cmidrule(lr){7-9}
      & 
      & \textbf{Latency ($\mathrm{s}$) $\downarrow$} 
      & \textbf{Speed $\uparrow$} 
      & \textbf{FLOPs ($\mathrm{T}$) $\downarrow$}  
      & \textbf{Speed $\uparrow$} 
      & \textbf{SC $\uparrow$} 
      & \textbf{PQ $\uparrow$} 
      & \textbf{OS $\uparrow$} \\
      \midrule

      FLUX.1-Kontext-dev 
      & 50 
      & 50.20 
      & 1.00$\times$ 
      & 8299.54 
      & 1.00$\times$ 
      & 6.80 
      & 7.26 
      & 6.51 \\
      \midrule

      SpecEdit($\tau=0.4$) 
      & 18 
      & 12.60 
      & 3.98$\times$ 
      & 1594.16 
      & 5.20$\times$ 
      & 6.99 
      & \textbf{7.12} 
      & 6.61 \\

      SpecEdit($\tau=0.90$) 
      & 18 
      & \textbf{12.20} 
      & \textbf{4.11$\times$} 
      & \textbf{1488.84} 
      & \textbf{5.57$\times$} 
      & 6.90 
      & 7.10 
      & 6.55 \\

      \rowcolor{gray!15}  
      \textbf{SpecEdit}($\tau=0.75$) 
      & 18 
      & 12.50 
      & 4.02$\times$ 
      & 1541.50 
      & 5.38$\times$ 
      & \textbf{7.12} 
      & 6.98 
      & \textbf{6.69} \\

      \bottomrule
    \end{tabular}
  }

  {\scriptsize
  \begin{itemize}[leftmargin=10pt,topsep=-2pt,label=\textbf{•}]
    \item \textbf{SC}: semantic consistency, \textbf{PQ}: perceptual quality, \textbf{OS}: overall score.
  \end{itemize}
  }
\end{center}
\paragraph{Effect of the verification threshold.}
Table~\ref{table:upsample_different_method} studies the effect of the discrepancy threshold $\tau$ on FLUX.1-Kontext-dev with GEdit-Bench. 
The results show that the threshold controls a clear trade-off between efficiency and editing fidelity. 
A smaller threshold such as $\tau=0.4$ selects more tokens for high-resolution refinement, which leads to better perceptual quality with PQ reaching \textbf{7.12}, but the computational cost is slightly higher. 
In contrast, a larger threshold such as $\tau=0.90$ is more aggressive in filtering tokens and therefore achieves the best efficiency with \textbf{4.11$\times$} latency speedup and \textbf{5.57$\times$} FLOPs reduction, but this comes with a drop in semantic consistency and overall score. 
Our default setting $\tau=0.75$ provides the best balance. It attains the highest overall score of \textbf{6.69} together with the strongest semantic consistency of \textbf{7.12}, while still maintaining strong acceleration at \textbf{4.02$\times$}. 
These results indicate that moderate semantic verification is sufficient to focus computation on truly edited regions while avoiding the instability caused by overly sparse refinement.

\begin{center}
  \setlength\tabcolsep{4pt}
  \small

  \captionof{table}{Ablation Study on FLUX.1-Kontext-dev with GEdit-Bench}
  \label{table:draft_length_ablation}

  \resizebox{\textwidth}{!}{
    \begin{tabular}{l c c c c c c c c}
      \toprule
      \multirow{2}{*}{\textbf{Method}} & \multirow{2}{*}{\textbf{NFE}} 
      & \multicolumn{4}{c}{\textbf{Acceleration}}
      & \multicolumn{3}{c}{\textbf{GEdit-EN(FULL)}} \\
      \cmidrule(lr){3-6} \cmidrule(lr){7-9}
      & & \textbf{Latency ($\mathrm{s}$) $\downarrow$} & \textbf{Speed $\uparrow$} 
      & \textbf{FLOPs ($\mathrm{T}$) $\downarrow$}  & \textbf{Speed $\uparrow$} 
      & \textbf{SC $\uparrow$} & \textbf{PQ $\uparrow$} & \textbf{OS $\uparrow$} \\
      \midrule

      FLUX.1-Kontext-dev 
      & 50 
      & 50.20 
      & 1.00$\times$ 
      & 8299.54 
      & 1.00$\times$ 
      & 6.80 
      & 7.26 
      & 6.51 \\
      \midrule

      SpecEdit($N_{\mathrm{draft}}=4$) 
      & 18 
      & \textbf{12.05} 
      & \textbf{4.17$\times$} 
      & \textbf{1488.08} 
      & \textbf{5.58$\times$} 
      & 7.01 
      & \textbf{7.10} 
      & 6.64 \\

      \rowcolor{gray!15}  
      \textbf{SpecEdit}($N_{\mathrm{draft}}=8$) 
      & 18 
      & 12.24 
      & 4.10$\times$ 
      & 1541.50 
      & 5.38$\times$ 
      & \textbf{7.12} 
      & 6.98 
      & \textbf{6.69} \\

      \bottomrule
    \end{tabular}
  }

  {\scriptsize
  \begin{itemize}[leftmargin=10pt,topsep=-2pt,label=\textbf{•}]
    \item \textbf{SC}: semantic consistency, \textbf{PQ}: perceptual quality, \textbf{OS}: overall score.
  \end{itemize}
  }
\end{center}

\paragraph{Effect of the draft sampling length.}
We further study the number of draft inference steps in the preliminary verification stage. 
As shown in Table~\ref{table:draft_length_ablation}, both settings achieve strong acceleration and competitive editing quality, which suggests that the draft stage is robust to moderate changes in sampling length. 
Using $N_{\text{draft}}=4$ yields slightly better efficiency with \textbf{4.17$\times$} latency speedup and \textbf{5.58$\times$} FLOPs reduction, and it also achieves the best perceptual quality of \textbf{7.10}. 
Using $N_{\text{draft}}=8$ produces the best semantic consistency and overall score, reaching \textbf{7.12} SC and \textbf{6.69} OS. 
This result suggests that a slightly longer draft trajectory provides a more reliable semantic estimate for token verification, which improves the final balance between semantic preservation and perceptual quality. 
Unless otherwise specified, we use $N_{\text{draft}}=8$ as the default setting in the main experiments.

\vspace{100pt}

\begin{center}
  \setlength{\tabcolsep}{2pt}  
  \footnotesize  
  \captionof{table}{Quantitative results of image-editing on ImgEdit-Bench.}
  \label{tab:imgedit_flux_ablation}  
  \resizebox{\linewidth}{!}{
  \begin{tabular}{l l >{\extracolsep{-3pt}}c c<{\extracolsep{0pt}} *{10}{c}} 
    \toprule  
    \multirow{2}{*}{\textbf{Model}} & 
    \multirow{2}{*}{\textbf{NFE}} & 
    \multicolumn{2}{c}{\textbf{Acceleration}} & 
    \multicolumn{10}{c}{\textbf{ImgEdit-Bench}} \\
    \cmidrule(lr){3-4} \cmidrule(lr){5-14}  
    & & \textbf{Latency ($\text{s}$) $\downarrow$} & \textbf{Speedup $\uparrow$} & 
    \textbf{Action} & \textbf{Add} & \textbf{Adjust} & \textbf{Background} & 
    \textbf{Compose} & \textbf{Extract} & \textbf{Remove} & \textbf{Replace} & 
    \textbf{Style} & \textbf{Overall $\uparrow$} \\
    \midrule  
    \textbf{$100\%$ steps}        & 50 & 50.20 & 1.00$\times$ & 3.89 & 4.41 & 3.60 & 4.05 & 3.24 & 4.00 & 3.47 & 3.20 & 2.84 & 3.66 \\ 

    \midrule

    FLUX.1-Kontext-Lightning         & 8 & 8.24 & 6.09$\times$ & \textbf{3.88} & 4.37 & 3.53 & 3.98 & \textbf{3.37} & \textbf{3.86} & \textbf{3.37} & 3.24 & \textbf{2.78} & \textbf{3.60}  \\

    FLUX.1-Kontext-Lightning         & 6 & 6.36 & 7.89$\times$ & 3.84 & 4.37 & 3.45 & \textbf{4.14} & 3.22 & 3.78 & 3.33 & 3.22 & 2.73 & 3.56 \\

    \rowcolor{gray!15}  
    \textbf{\textbf{SpecEdit}+Step Distillation}        & 8 & 6.97 & 7.20$\times$ & 3.87 & 4.37 &3.54  & 4.04 &3.33  & 3.83 & 3.19 & \textbf{3.27} & 2.72 & 3.57 \\
    \rowcolor{gray!15}  
    \textbf{\textbf{SpecEdit}+Step Distillation}        & 6 & \textbf{5.48} & \textbf{9.16$\times$} & 3.83 & \textbf{4.39} & \textbf{3.58} & 4.10 & 3.30 & 3.75 & 3.25 & 3.23 & 2.68 & 3.57 \\

    \bottomrule  
  \end{tabular}}
\end{center}

\paragraph{Compatibility with step distillation.}
Table~\ref{tab:imgedit_flux_ablation} shows that SpecEdit is naturally compatible with step distillation on FLUX.1-Kontext-dev. 
When combined with FLUX.1-Kontext-Lightning at 8 NFE, SpecEdit improves the latency from 8.24s to 6.97s, increasing the speedup from \textbf{6.09$\times$} to \textbf{7.20$\times$}, while maintaining a very similar overall score of 3.57 compared with 3.60 for the distilled baseline. 
At 6 NFE, the gain becomes more significant. SpecEdit further reduces the latency to \textbf{5.48s} and reaches \textbf{9.16$\times$} speedup, while preserving the same overall score of 3.57. 
It also improves several category scores including \textit{Add} and \textit{Adjust}, which indicates that semantics-aware spatial allocation complements temporal compression effectively. 
These results confirm that SpecEdit is orthogonal to distillation based acceleration and can provide additional efficiency gains without noticeably degrading editing quality.

\section{Supplementary Visualizations}
\begin{figure*}[htbp]
  \centering
  \includegraphics[width=1\linewidth]{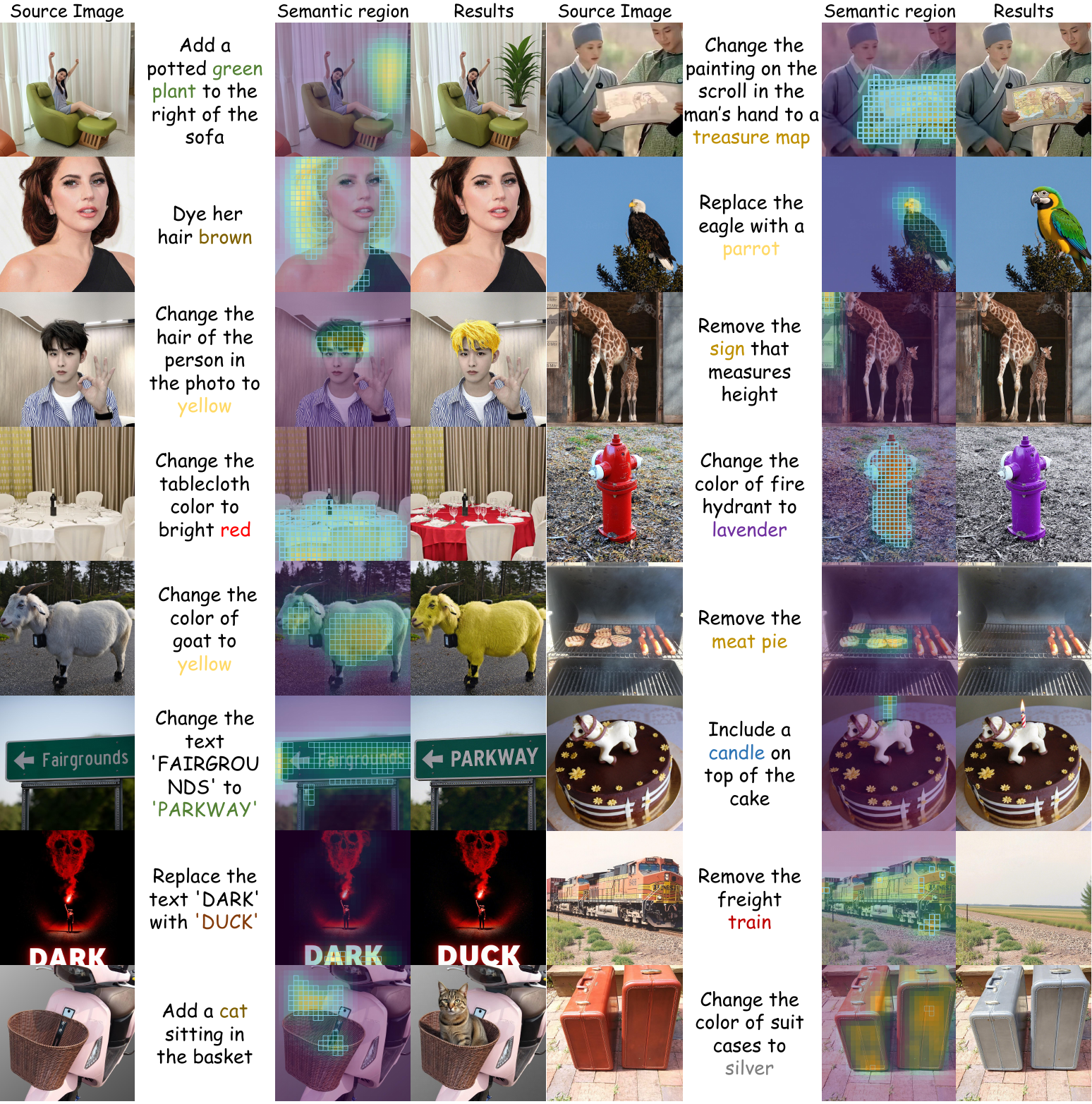}
  \caption{ More visualization of semantic locking captured by SpecEdit. }
  \label{fig:visualization}
\end{figure*}

\begin{figure*}[htbp]
  \centering
  \includegraphics[width=1\linewidth]{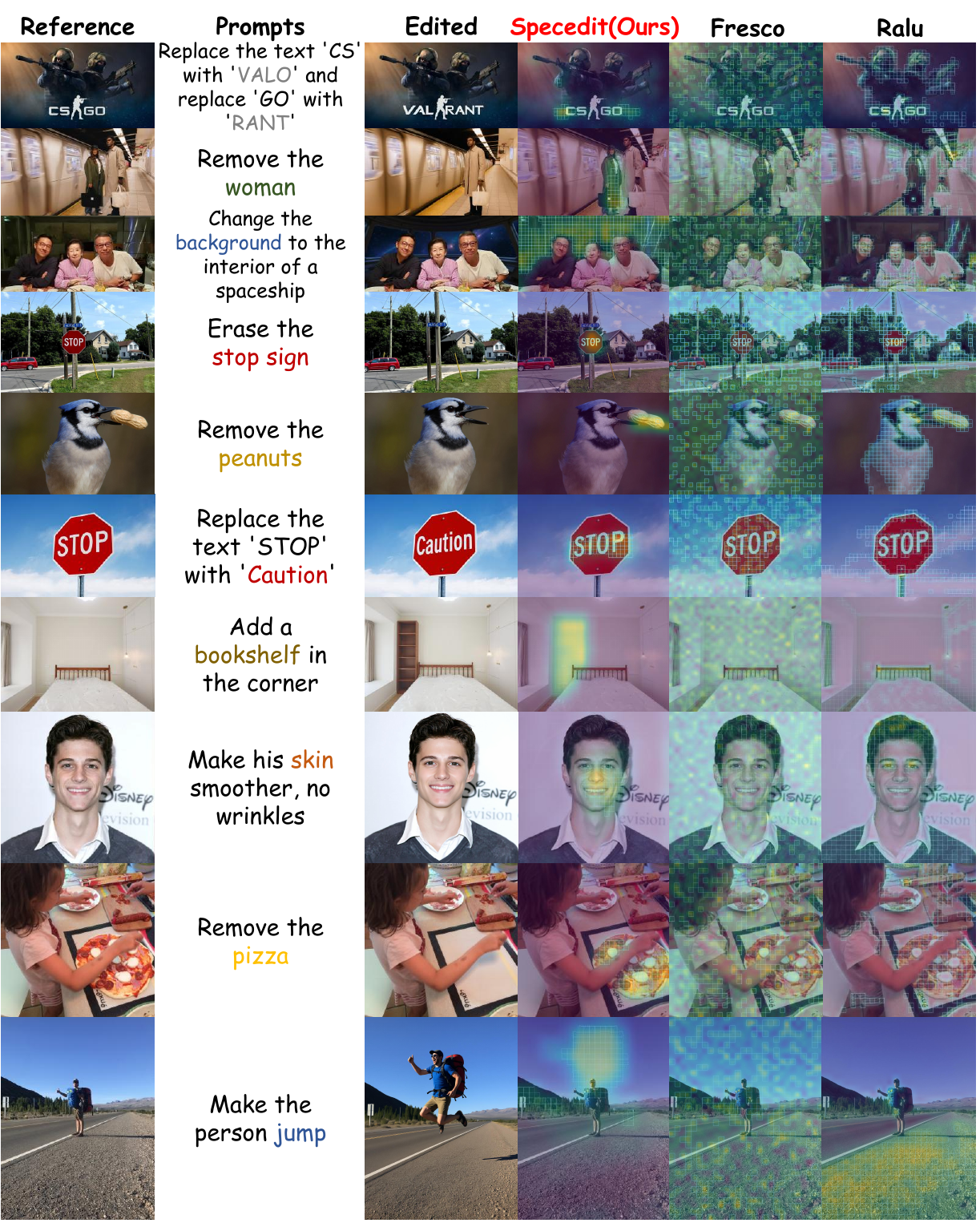}
  \caption{
  \textbf{Comparison of partially upsampled regions.}
  We visualize the spatial regions selected for high-resolution refinement by different methods, including \textbf{RALU}, \textbf{Fresco}, and \textbf{SpecEdit} (ours).
  RALU and Fresco rely on low-level structural cues such as edges or channel statistics, which often activate regions unrelated to the editing instruction.
  In contrast, \textbf{SpecEdit} identifies semantically relevant areas through draft-based verification, resulting in more accurate localization of edited regions and fewer redundant high-resolution computations.
  }
  \label{fig:Comparison-upsampled}
\end{figure*}

\paragraph{Overview of supplementary visualizations.}
Figure~\ref{fig:downsampling} analyzes the effect of different downsampling ratios in the draft stage and shows that $16\times$ downsampling preserves sufficient structural information for reliable localization of edited regions, while more aggressive downsampling removes important details and leads to inaccurate identification of edited areas.
Figure~\ref{fig:visualization} visualizes the semantic locking effect captured by SpecEdit, highlighting how the method concentrates high-resolution computation on semantically modified regions.
Figure~\ref{fig:Comparison-upsampled} compares the partially upsampled regions selected by different methods, showing that SpecEdit identifies editing-relevant areas more accurately than heuristic approaches such as RALU and Fresco and avoids unnecessary refinement of irrelevant regions.
Figure~\ref{fig:specedit_gedit_compare} presents additional qualitative comparisons on GEdit-Bench, demonstrating that SpecEdit maintains better semantic consistency and visual quality under high acceleration compared with existing acceleration methods.

\begin{figure*}[htbp]
  \centering
  \includegraphics[width=1\linewidth]{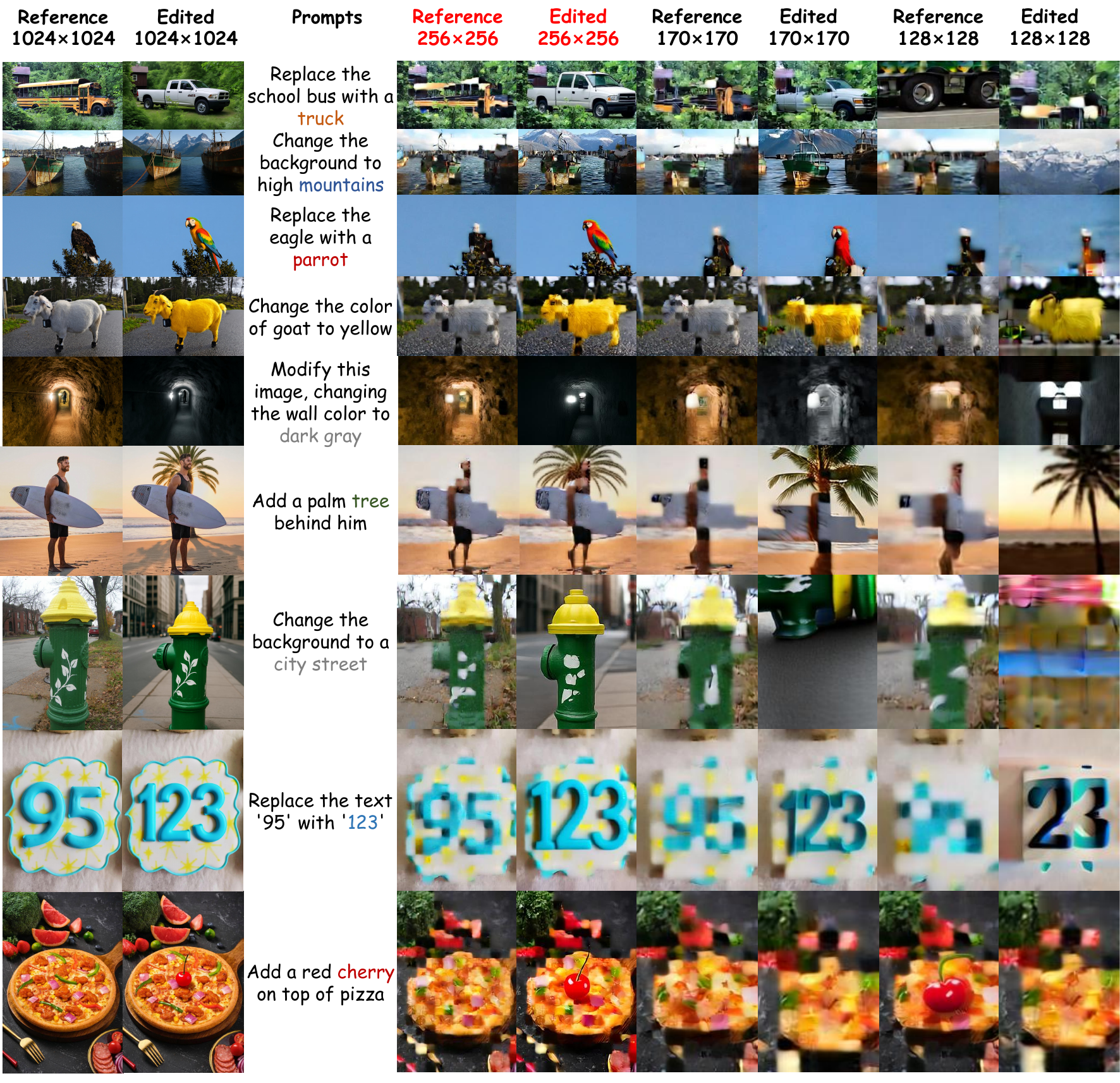}
  \caption{
  \textbf{Effect of different downsampling ratios on preserving editable regions.}
  We compare the reference and edited images at the original resolution (1024$\times$1024) together with their downsampled versions (256$\times$256, 170$\times$170, and 128$\times$128), corresponding to \textbf{16$\times$, 36$\times$, and 64$\times$ downsampling}. 
  When using \textbf{16$\times$ downsampling}, most structural and semantic information of the image is preserved, enabling the edited regions to be reliably localized by comparing the reference and denoised images. 
  In contrast, more aggressive downsampling (\textbf{36$\times$} and \textbf{64$\times$}) removes substantial spatial details, making it difficult to accurately identify the modified areas.
  }
  \label{fig:downsampling}
\end{figure*}

\begin{figure*}[htbp]
  \centering
  \includegraphics[width=1\linewidth]{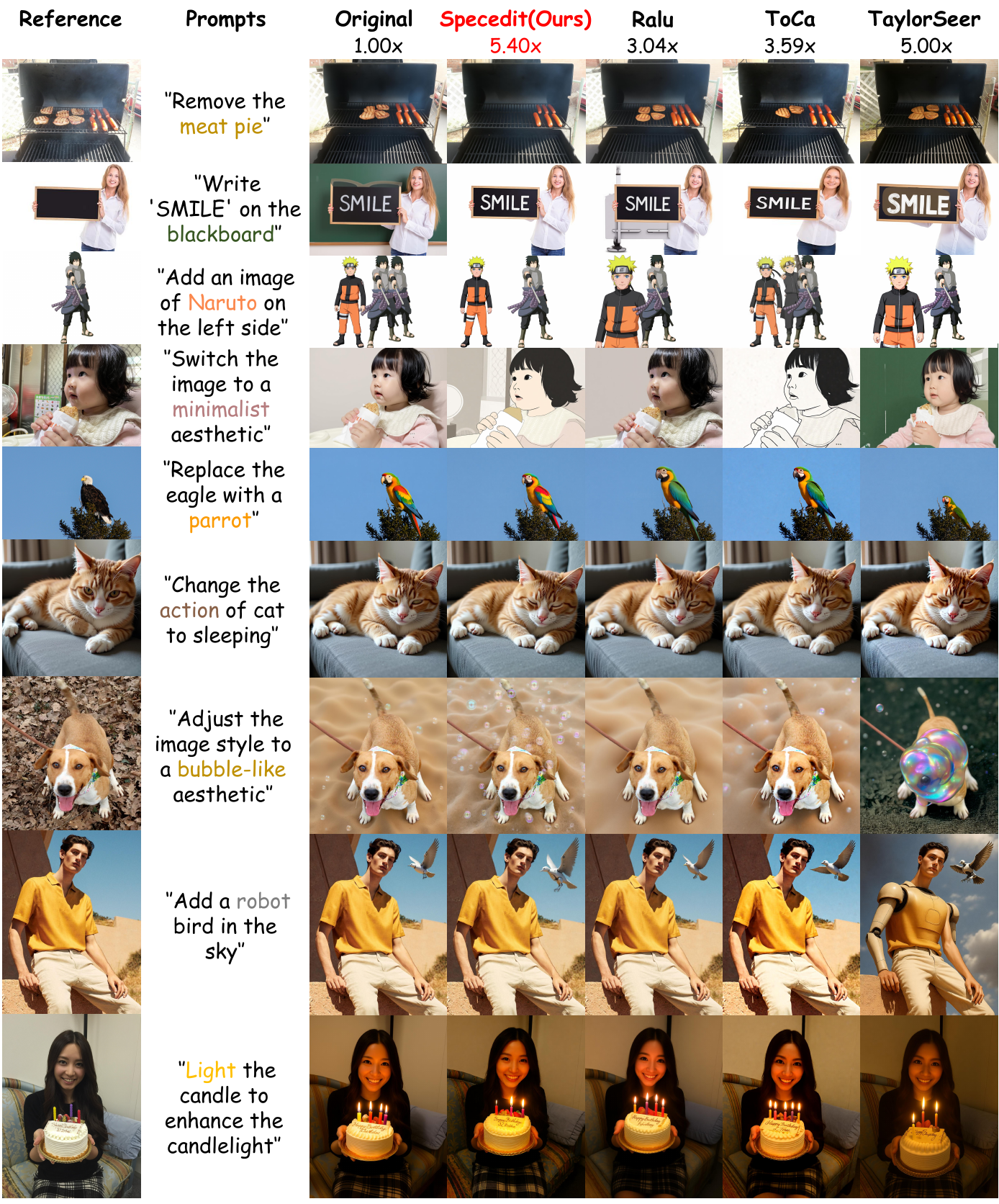}
  \caption{ \textbf{More qualitative comparison on GEdit-Bench.} SpecEdit achieves superior semantic preservation and visual quality at a higher acceleration ratio (\textbf{5.38$\times$}) compared to state-of-the-art baselines. }
  \label{fig:specedit_gedit_compare}
\end{figure*}

\end{document}